\pdfoutput=1

\documentclass[11pt]{article}

\usepackage[preprint]{acl}

\usepackage{times}
\usepackage{latexsym}
\usepackage{algorithm}
\usepackage{algpseudocode}
\usepackage{amsmath}
\usepackage{booktabs,siunitx}
\usepackage{tikz}
\usepackage{adjustbox}
\usepackage{multirow}
\usepackage{multicol}
\usepackage{arydshln}
\usepackage{array}
\usepackage{float}
\usepackage[normalem]{ulem}
\usepackage{amssymb}    
\usepackage{xcolor}     
\newcommand{\redcross}{\textcolor{red}{\texttimes}}

\definecolor{fgreen}{HTML}{228B22}

\definecolor{blue1}{HTML}{001A6E}
\definecolor{blue2}{HTML}{074799}
\definecolor{blue3}{HTML}{009990}

\usepackage[T1]{fontenc}

\usepackage[utf8]{inputenc}

\usepackage{microtype}

\usepackage{inconsolata}

\usepackage{graphicx}
\usepackage{relsize}
\usepackage{anyfontsize}
\usepackage[breakable]{tcolorbox}

\definecolor{deepgreen}{rgb}{0.0, 0.5, 0.0}

\newcommand{\greencheck}{}%
\DeclareRobustCommand{\greencheck}{%
  \tikz\fill[scale=0.4, color=deepgreen]
  (0,.35) -- (.25,0) -- (1,.7) -- (.25,.15) -- cycle;%
}

\title{Context-Aware Hierarchical Merging for Long Document Summarization}

\author{Litu Ou \\
  University of Edinburgh \\
  \texttt{lituou907@gmail.com} \\\And
  Mirella Lapata \\
  University of Edinburgh \\
  \texttt{mlap@inf.ed.ac.uk} \\}

\begin{document}
\maketitle
\begin{abstract}
Hierarchical Merging  is a technique commonly used to summarize very long texts ($>$100K tokens) by breaking down the input into smaller sections, summarizing those sections individually, and then merging or combining those summaries into a final coherent summary. Although it  helps address the limitations of large language models (LLMs) with fixed input length constraints, the recursive merging process can amplify LLM hallucinations, increasing the   risk of factual inaccuracies. In this paper, we seek to mitigate hallucinations by enriching hierarchical merging with  context from the source document. Specifically, we propose different approaches to contextual augmentation ranging from \emph{replacing} intermediate summaries with relevant input context, to \emph{refining} them while using the context as supporting evidence, and \emph{aligning} them implicitly (via citations) to the input.   
Experimental results on datasets representing legal and narrative domains show that contextual augmentation consistently outperforms zero-shot and hierarchical merging baselines for the Llama 3.1 model family. Our analysis further reveals that refinement methods tend  to perform best when paired with extractive summarization for identifying relevant input\footnote{Our code and data are available at: \url{https://github.com/Leonard907/CAHM}}.

\end{abstract}

\section{Introduction}

\begin{figure}[t]
  \includegraphics[width=0.46\textwidth, trim=0cm 0.2cm 0cm 0cm, clip]{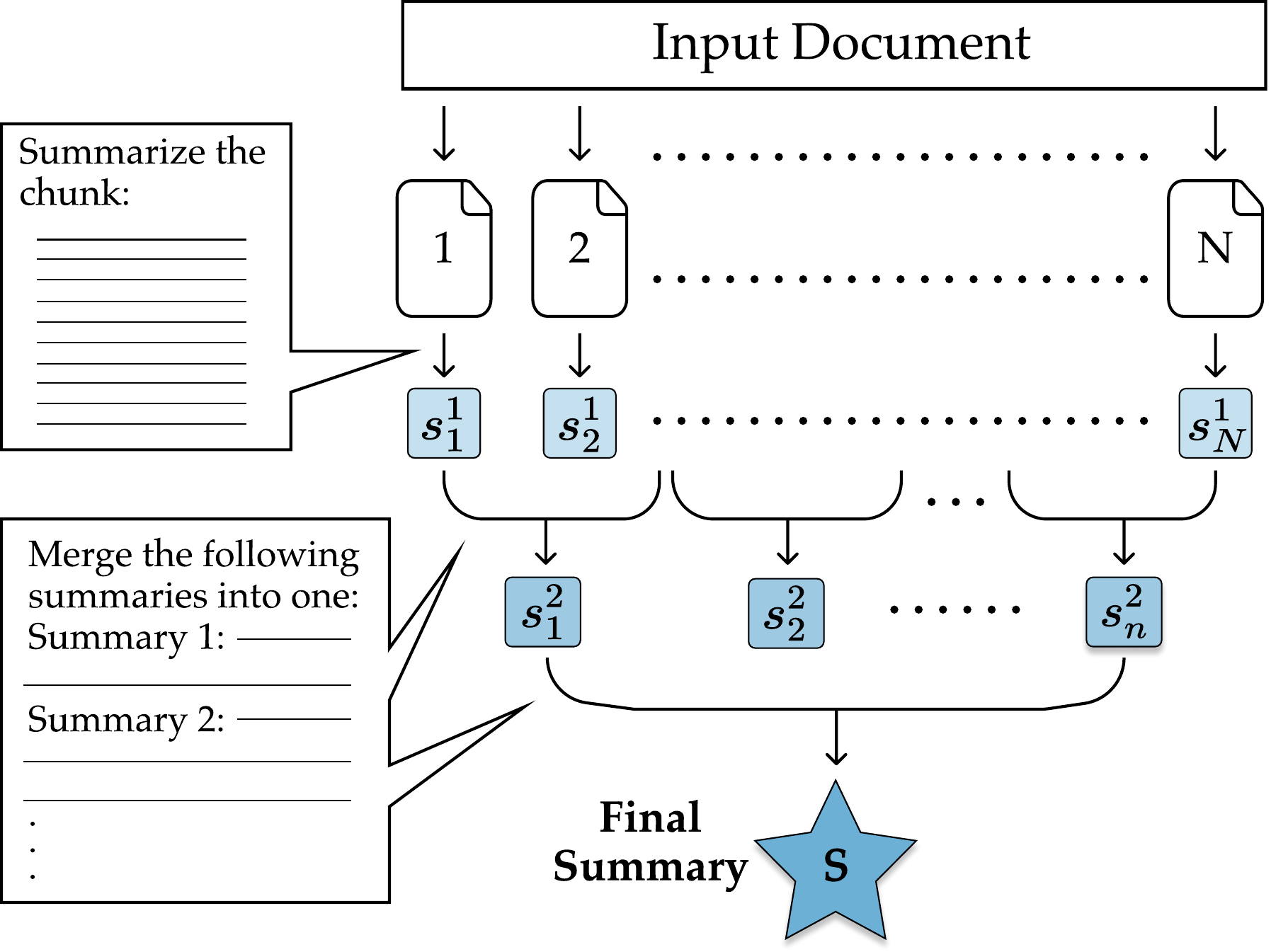}
  \caption{Illustration of hierarchical merging. The document is split into chunks and a first-level summary~$s^1_i$ is generated for each.  Second-level summaries~$s^2_j$ are created by merging  first-level summaries, and the final summary is produced by merging second-level summaries. Message boxes on the left provide examples of the prompts used for summarizing chunks and merging summaries at each level.}
  \label{fig:hm}
\end{figure}

Large language models (LLMs) have made significant strides in processing longer contexts, with models like Gemini \cite{Reid2024Gemini1U} reportedly being able to handle up to 2~million tokens. Despite promising potential for complex real-world applications such as  long-document question-answering and repository-level code understanding, 
most LLMs, especially open-source ones, struggle with tasks operating over long input sequences \cite{andryushchenko2024leveraginglargelanguagemodels}. A notable application we focus on in this paper is the summarization of very long documents  in  domains such as law and narrative which typically exceed  input size limitations of most current models ($>$100K tokens). 



Previous studies have  introduced specialized techniques for summarizing very long texts, with  \textit{hierarchical merging} \cite{Wu2021RecursivelySB, chang2024booookscore} being  one of the most popular methods. As illustrated in Figure~\ref{fig:hm},  hierarchical merging first divides an input document  into chunks of fixed token length, and a summary is generated for each chunk. In subsequent stages, consecutive chunk-level summaries are concatenated and merged until they reach a pre-defined length limit, after which a new summary is generated from the merged text. Merging continues until a single final summary is produced. All summary generation steps are performed using LLMs and  zero-shot prompting. 

Although hierarchical merging effectively handles inputs exceeding model length limits, it cannot produce faithful summaries even when using the most performant LLMs \cite{fables-2024-kim-et-al,li2024longcontextllmsstrugglelong}. In this paper, we seek to mitigate hallucinations and enhance summary faithfulness by enriching hierarchical merging with \emph{relevant context} from the source document. 
Access to input information during the merging process should provide  models with \emph{factual information} to generate more accurate summaries. 
A key question concerns how relevant context can be identified and used in the merging process. We propose three methods for selecting  context from the source document: 1)~\emph{extractive summarization} identifies key sentences from the input to produce a coherent abstractive summary; 2)~\emph{retrieval}  uses initial summaries as queries to retrieve relevant input passages, and 3)~\emph{attribution} generates summaries with references to input passages. Once relevant context is identified, it can either \emph{replace} the abstractive summary or serve as \emph{supporting evidence} for subsequent merging steps.

Experimental results with the Llama-3.1 model family, demonstrate
consistent  improvements (across  metrics) on summary faithfulness and quality, compared to the original hierarchical merging method and zero-shot summarization. Our analysis further  indicates that abstractive summaries play a crucial role in hierarchical merging by ensuring comprehensiveness, while augmenting them with relevant  contexts enhances summary faithfulness. Our contributions are summarized as follows:
\begin{itemize}
\item We develop a novel pipeline that incorporates input context into hierarchical merging for very long document summarization.
\item We propose three methods for selecting contexts: extractive summarization, retrieval, and text generation with citations.
\item We analyze  how the mechanism of contextual augmentation as a replacement  to intermediate summaries or supporting evidence affects the faithfulness of the final summary.
\end{itemize}

\section{Related Work}
\label{sec:related_work}


The literature is rife with proposals aiming to  mitigate LLM difficulties with processing inputs that exceed their context lengths. A significant challenge in overcoming the context limit is addressing the quadratic computational
burden of the self-attention mechanism. Previous work has 
attempted to reduce this cost by architectural modifications such as introducing sparse attention \cite{child2019generatinglongsequencessparse,beltagy2020longformerlongdocumenttransformer}, linearized attention \cite{katharopoulos:ea:2020},  changes to positional encoding \cite{peng2023yarnefficientcontextwindow,chen2023extendingcontextwindowlarge},  and special-purpose fine-tuning methods which directly modify the attention mechanism \cite{chen2024longloraefficientfinetuninglongcontext}  or incorporate discourse-level information. 

Another line of work adopts a divide-and-conquer approach, splitting the long input into manageable chunks. The chunks can be processed independently \cite{wang2023augmentinglanguagemodelslongterm,bertsch2023unlimiformer} or through progressively merging adjacent chunks as they
are processed along the transformer layers \cite{song2024hierarchicalcontextmergingbetter}. 
\citet{Wu2021RecursivelySB} propose a dive-and-conquer approach for long document summarization in which an LLM is fine-tuned via reinforcement learning to summarize each chunk and then hierarchically merge
chunk-level summaries  into a final summary. Their method has since been adapted to use zero-shot prompting for summary generation \cite{chang2024booookscore, fables-2024-kim-et-al}, without further training (as shown in Figure~\ref{fig:hm}). 

Our proposal modifies hierarchical merging to consider evidence from the source document at the intermediate summary generation stage. It draws inspiration from retrieval augmented generation (RAG) techniques which have proven particularly effective in long-form question answering \cite{lewis2020rag, izacard-grave-2021-leveraging,xu2024retrievalmeetslongcontext,edge2024localglobalgraphrag}, as a means to integrate up-to-date information, mitigate hallucinations, and enhance response quality. Our work explores various proposals for context integration, e.g.,~as a replacement of  intermediate summaries or additional supporting evidence, as well as retrieval augmentation methods. In addition to traditional query-based retrieval which we implement using the intermediate summaries as queries, we also turn to extractive summarization to select and rank source document sentences that best represent its content.  

Various methods have been proposed for long document extractive
summarization, focusing on reinforcement learning
\cite{gu-etal-2022-memsum, bian2023gosumextractivesummarizationlong}
and adapting LLMs for sentence extraction \cite{Lu2023HybridLD,
  Hemamou2024ScalingUS}. However, these methods were primarily
evaluated on datasets with relatively short inputs, such as arXiv
(\citealt{cohan-etal-2018-discourse}; ~5K tokens) and GovReport
(\citealt{huang-etal-2021-govreport}; ~10K tokens), making their
effectiveness on inputs exceeding 100K tokens uncertain. As we
identify salient sentences within each chunk, we expect extractive
methods to be able to isolate supporting evidence for ultimately
generating more factual \emph{abstractive} summaries.

Our work also  has connections to recent efforts in creating verifiable systems that generate responses to queries by incorporating citations to source material \cite{fierro2024learningplangeneratetext,nakano2022webgptbrowserassistedquestionansweringhuman}. Most existing models learn to generate citations \cite{menick2022teachinglanguagemodelssupport,nakano2022webgptbrowserassistedquestionansweringhuman} or implement a post-processing step \cite{Bohnet:Ea:2022,gao-etal-2023-enabling} where the text is first generated and then edited to be made attributable (e.g.,~to retrieved web content).  For hierarchical merging, generating intermediate summaries with citations is advantageous as it  eschews the need for additional retrieval or extraction mechanisms. However, we experimentally find that generating \emph{accurate} citations is challenging requiring models to understand and execute complex instructions. 




\begin{figure}[t]
  \includegraphics[width=0.45\textwidth, trim=0cm 0cm 0cm 0cm, clip]{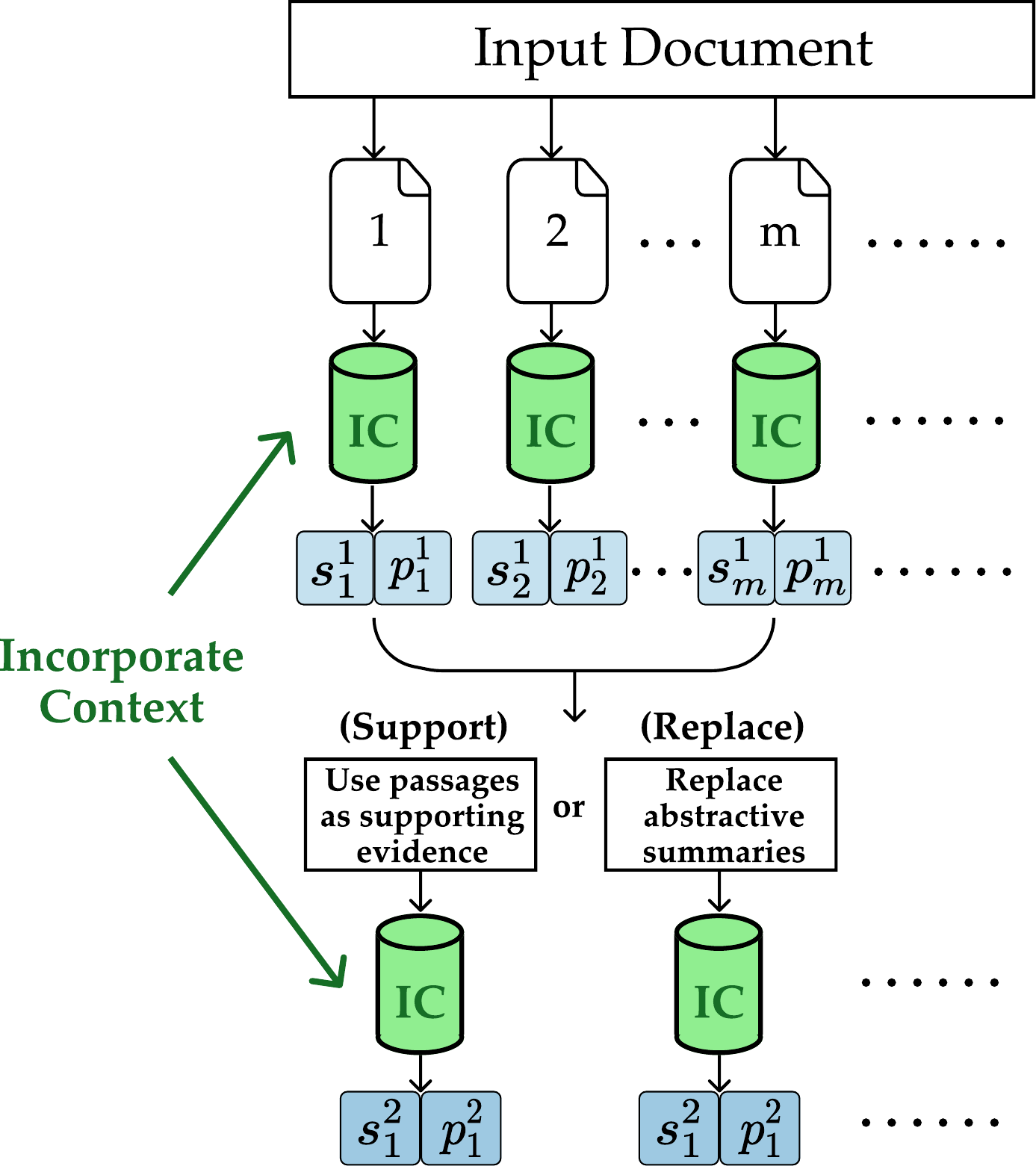}
  \caption{Illustration of proposed pipeline for the first two
    levels. The Incorporate Context (IC) module is applied at the
    first-level to obtain relevant input contexts~$p^1_i$ alongside
    summaries~$s^1_i$. At the \mbox{second-level}, we either use
    $p^1_1 \dots p^1_m$ as supporting evidence (\textbf{Support}), or
    replace $s^1_1 \dots s^1_m$ entirely with $p^1_1 \dots p^1_m$
    (\textbf{Replace}) for generating~$s^2_1$ and subsequently
    obtaining~$p^2_1$.}
  \label{fig:proposed}
\end{figure}

\begin{figure*}[t]
  \includegraphics[width=\textwidth, trim=0cm 0cm 0cm 0cm, clip]{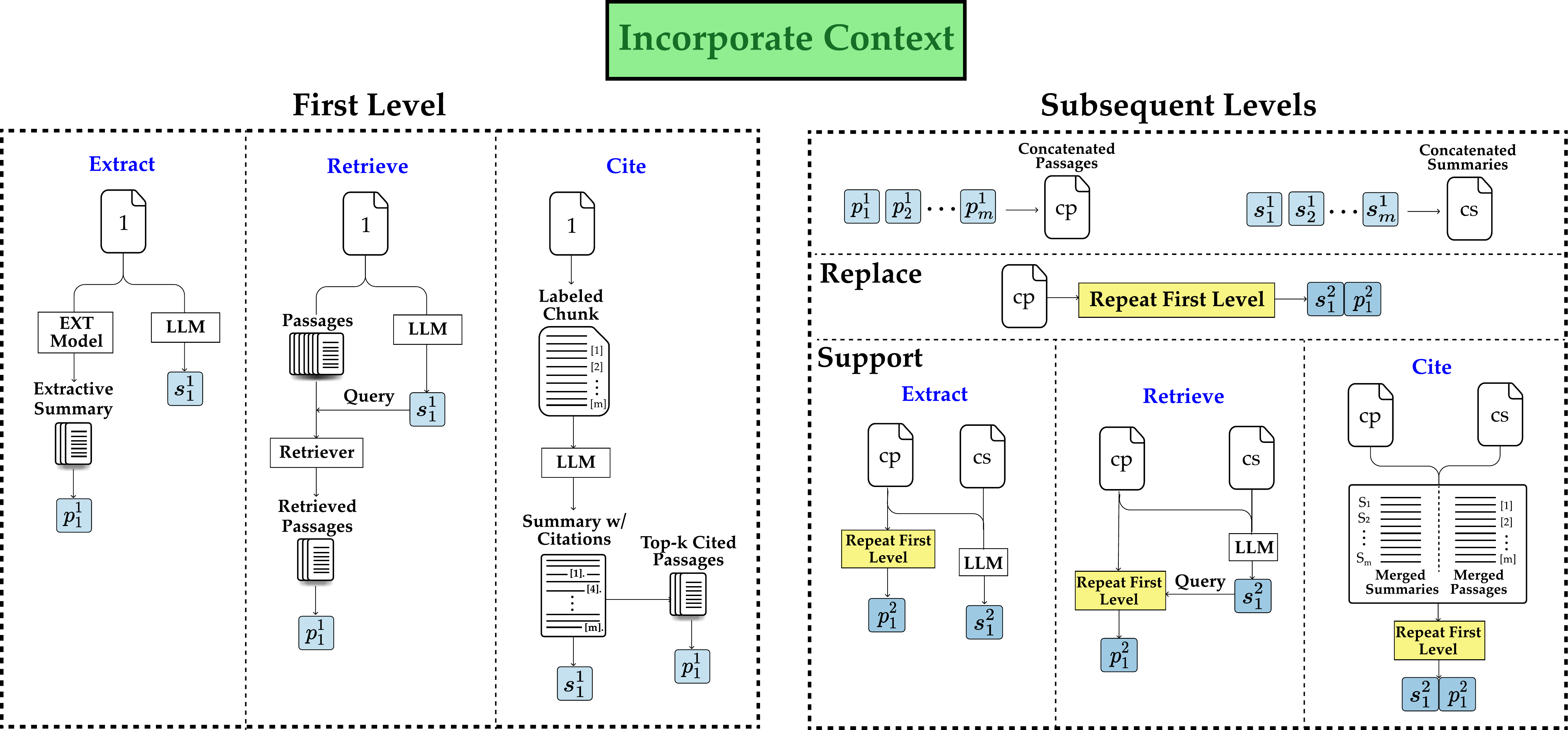}
 \caption{\label{fig:ic} Incorporate Context (IC)
   module from Figure~\ref{fig:proposed} with three context
   augmentation methods: Extractive Summarization (Extract),
   Retrieval-augmented Generation (Retrieve), and Generating text with
   citations (Cite).} 
\end{figure*}

\section{Hierarchical Merging with Context}
\label{sec:methods}

Figure~\ref{fig:proposed} illustrates our proposed hierarchical merging pipeline for the first two levels. Compared to the original method in Figure~\ref{fig:hm}, at each level, an additional Incorporate Context (IC) module is used to obtain relevant contexts alongside abstractive summaries. When generating  intermediate summaries for the current level, we either use the  summaries from the previous level \emph{and} relevant context as supporting evidence  (denoted as Support), or replace  the summaries with their corresponding contexts (denoted as Replace). At each level, the IC module takes abstractive summaries  and input contexts from the previous level to obtain relevant  contexts for the current level.  We make sure that input contexts have similar length to their respective abstractive summaries, so that the former can serve as a substitute for the latter. 

Figure~\ref{fig:ic} provides an overview of the IC module. At the first-level, chunks from the input document are used to obtain abstractive summaries and corresponding input contexts. In Sections~\mbox{\ref{sec:ic_ext}--\ref{sec:ic_cite}}, we explain in detail how we propose to obtain relevant contexts. 
In subsequent levels, merging involves
abstractive summaries (Concatenated Summaries) and their   contexts (Concatenated Passages).  If we choose to Replace summaries with their context, we simply substitute chunks with Concatenated Passages and repeat the first-level process. If we use input contexts as supporting evidence (Support), next level contexts are selected from Concatenated Passages, while next level summaries are generated by leveraging Concatenated Passages and Concatenated Summaries. Prompts for summary generation and merging are given in Appendix~\ref{sec:prompts}.

To obtain contexts that are as relevant as possible, we utilize and
modify three different methods which we describe as follows. The
implementation of these methods in our experiments is further
explained in Section~\ref{sec:experiment_setup}.

\subsection{Extractive Summarization: Extract}
\label{sec:ic_ext}

Extractive summarization selects salient sentences from the source
document to form a summary. Since extractive summarizers are trained
using abstractive summaries as targets
\cite{liu-lapata-2019-text,narayan-etal-2018-ranking,xu2022textsummarizationoracleexpectation},
they naturally select sentences that cover key information similar to their corresponding abstractive summaries, making them ideal input contexts for either replacement or supporting evidence. At the
first-level, the IC module applies extractive summarization to input chunks.  At subsequent levels, extraction operates over concatenated passages from the previous-level, ensuring selected contexts always
originate from the source document.

\subsection{Information Retrieval: Retrieve}
\label{sec:ic_retrieve}
Our use of information retrieval is motivated by the recent success
of retrieval-augmented language models at minimizing hallucinations
and improving the factuality of generated outputs
\cite{shi2023replugretrievalaugmentedblackboxlanguage,asai2023selfraglearningretrievegenerate,asai2024reliableadaptableattributablelanguage}. In the context of our summarization task, we use
intermediate summaries as queries to retrieve relevant input
contexts. Compared to Extract, this approach leverages the ability of
recent LLMs to distill important information from the input and
describe it succinctly. At the first-level, the IC module generates an
abstractive summary for each chunk. We then split each chunk into
passages of equal length (\textasciitilde100 words) and use the
abstractive summary as query to retrieve top-$k$ passages from the
chunk. At subsequent levels, the retriever selects input contexts from
concatenated passages at the previous-level.

\subsection{Text Generation with Citations: Cite}
\label{sec:ic_cite}
The generation of intermediate summaries with citations as supporting evidence to input passages
\cite{nakano2022webgptbrowserassistedquestionansweringhuman,fierro2024learningplangeneratetext}, avoids the use of
information retrieval. We split the input into passages of equal
length like in Section~\ref{sec:ic_retrieve}, and provide LLMs with explicit
instructions to cite  passages after each sentence in
their summaries (see Appendix~\ref{sec:prompts}, prompts in Tables~\ref{tab:prompts_first_lvl} and~\ref{tab:prompt_sub_level_support}). As a result, the summaries  show (via their citations) which passages are used to generate which
sentence. We then extract citations from the summary and
rank them based on citation frequency, selecting the top-$k$
passages as input contexts. When passages share equal citation counts,
we prioritize those that enhance coverage across different input
sections. Details of
this selection algorithm are given in Appendix~\ref{sec:alg_cite}.

At the first-level, we add a label after each passage in square
brackets with ascending numbers ($[1] \cdots [m]$; see Figure~\ref{fig:ic}). A summary with
citations is then generated from the labeled chunks. We employ the
selection algorithm in Appendix~\ref{sec:alg_cite} to identify relevant
contexts, before stripping the citations to get a clean summary that
will be used later. At subsequent levels, we consider two cases:

\begin{itemize}
    \item Replace: We repeat the first-level process, with 
      concatenated passages as  input.
    \item Support: We instruct the LLM to generate a summary based on
       concatenated summaries from the previous level and 
      concatenated passages   the citations are pointing to. 
\end{itemize}

\section{Experimental Setup}
\label{sec:experiment_setup}

\paragraph{Datasets} Our experiments are conducted on two long
document summarization datasets. The first dataset, Multi-LexSum
\cite{Shen2022MultiLexSum}, comprises U.S. federal court documents
paired with expert-written summaries at multiple levels of granularity
(long, short, and tiny summaries). For evaluation purposes, we utilize
the long summaries as a reference and exclude documents with fewer
than 100K~tokens. The second dataset, SuperSummary, is a novel book
summarization corpus that we constructed using literary works
published after~2022. The corresponding expert-written plot summaries
were obtained from
SuperSummary\footnote{https://www.supersummary.com/} through a paid
subscription service. Similarly, we employ these plot summaries as
gold summaries for evaluation and exclude documents containing fewer
than 100K~tokens. Table~\ref{tab:dataset_stat} presents the
statistics of both datasets.

\paragraph{Implementation Details} Our experimental evaluation was
conducted using the Llama 3.1 model family
\cite{dubey2024llama3herdmodels}, which features a context length of
128K tokens and comprises two variants with 8B and 70B parameters,
respectively. We use the GPTQ-INT4 quantized 70B model
\cite{frantar2023gptqaccurateposttrainingquantization} in our
experiments for inference efficiency. For all hierarchical merging
systems, the chunk size and maximum merging context length were
configured to 8K tokens, ensuring sufficient merging operations for
meaningful analysis of different hierarchical merging
variants.

As mentioned earlier, we propose three methods for incorporating
context, i.e.,~Extract, Retrieve, and Cite.  We use MemSum
\cite{gu-etal-2022-memsum}, a state-of-the-art extractive summarizer
trained with reinforcement learning, as our backbone model for
Extract. We fine-tune MemSum on the official Multi-LexSum training
split. For SuperSummary, we use the training split of the
publicly accessible BOOKSUM dataset
\cite{kryscinski-etal-2022-booksum}. To prepare these datasets for
MemSum fine-tuning, each document was segmented into \mbox{8K-token}
chunks, from which 1,000 chunks (per dataset) were randomly selected
as training inputs. Subsequently, we use the full Llama-3.1 70B model
to generate summaries for each chunk via zero-shot prompting, which
serve as reference summaries. The fine-tuning process  was
executed for 10 epochs with a learning rate of 1e-4. To maintain
consistency with the average summary length in the evaluation
datasets, the maximum number of extracted sentences was set to~20.

For Retrieve, we split input documents into uniform length passages
with~100 tokens each. To optimize compute efficiency, we utilize the
BM25 retriever to select passages using intermediate summaries as
queries. The top-$k$ parameter for selecting passages is set so as to
ensure concatenated top-$k$ passages have  lengths similar to the
average summary length in the evaluation datasets.  Similar to
Retrieve, we set the passage length to 100 tokens each for Cite. The
top-$k$ passage selection process is described in
Section~\ref{sec:alg_cite}.

\begin{table}[t]
  \centering \footnotesize \def\arraystretch{1.5}
  \begin{tabular}{lrr}
      \hline
       & \multicolumn{1}{c}{Multi-LexSum} & \multicolumn{1}{c}{SuperSummary} \\
      \hline
    
      \# examples & 151 & 194 \\
      input tokens & 156,447 & 133,246 \\
      summary tokens & 1,147 & 1,180 \\
       domain & legal  & narrative \\
      
      \hline
  \end{tabular}
  \caption{Statistics of the datasets used in this paper. We report the number (\#) of examples per dataset, and the average number of input and summary tokens.}
  \label{tab:dataset_stat}
\end{table}

\paragraph{Model Comparisons} As described in
Section~\ref{sec:methods}, the proposed hierarchical merging pipeline
offers two approaches for incorporating context beyond the first level:
either via replacing abstractive summaries with corresponding input
contexts (Replace) or via using input contexts as supporting evidence
(Support). This together with the three context selection methods
(Extract, Retrieve, Cite) yields six distinct model invitations:
Extract-Replace, Extract-Support, Retrieve-Replace,
Retrieve-Support, Cite-Replace, and Cite-Support.

We evaluate these models against three baseline approaches:
(a)~Zero-shot summarization (Zero-shot), which generates summaries
through zero-shot prompting after truncating the input to the maximum
context length; (b)~the original hierarchical merging method (HMerge),
as illustrated in Figure~\ref{fig:hm}; and an adaptation thereof  which incorporates citation
generation (Cite-HMerge). Each intermediate summary is stripped off its
citations and re-labeled together with other merged summaries at the
merging step.

\paragraph{Evaluation Metrics} We use three automatic metrics to evaluate the \emph{faithfulness} of our summaries: 

\textsc{SummaC} \cite{laban2021summacrevisitingnlibasedmodels} is a specialized metric developed for assessing the semantic consistency between generated summaries and input contexts. Its architecture is particularly effective for long-document summarization, as it implements natural language inference at the sentence level and employs a convolutional layer to aggregate results into a single final consistency score.

AlignScore \cite{zha-etal-2023-alignscore} evaluates the information alignment between generated texts and the corresponding source texts through a unified text-to-text alignment function. While \textsc{SummaC} is specifically optimized for summarization tasks, AlignScore is a more general metric and demonstrates robust performance across various applications, including summarization and fact verification. This metric implements chunk-based processing to handle long documents, similar to \textsc{SummaC}'s approach.

\textsc{PRisma} \cite{mahon-lapata-2024-modular}, is a recent summarization metric built upon \textit{FActScore} \cite{min-etal-2023-factscore}. This metric decomposes generated texts into atomic facts—discrete statements containing singular pieces of information—and evaluates factual consistency between generated and reference summaries at this granular level. The methodology involves a two-step process: first, extracting atomic facts from both generated and reference summaries, then computing fact precision by computing the number of generated summary facts entailed by the reference summary, and fact recall by computing the number of reference summary facts entailed by the generated summary. The final metric is derived from the harmonic mean of these precision and recall measurements. Unlike \textsc{SummaC} and AlignScore, \textsc{PRisma} evaluates summary quality through comparison with reference summaries, allowing it to better capture whether the generated summary preserves the same key information that human annotators deemed important in their reference summaries. We use the GPT-4o-mini model for both fact extraction and entailment computation, incorporating in-context learning examples to ensure the quality of atomic fact extraction, following \citet{mahon-lapata-2024-modular}.

In addition to faithfulness metrics, we report two standard
metrics that assess the \emph{quality} of generated summaries
against reference summaries: ROUGE \cite{lin-2004-rouge} as the
geometric mean of ROUGE-1, \mbox{ROUGE-2}, and ROUGE-L
\cite{shaham-etal-2023-zeroscrolls} and BERTScore
\cite{he2021debertadecodingenhancedbertdisentangled} (we use the
 \texttt{deberta-xlarge-mnli} model and report F1). Overall, we employ three input-based metrics (\textsc{SummaC} and
AlignScore), which evaluate against the source text and three
reference-based metrics (\textsc{PRisma}, ROUGE, and BERTScore), which
compare against reference summaries.

\section{Results}
\label{sec:results}

Tables \ref{tab:8b} and \ref{tab:70b} present our results for the Llama-3.1 8B and 70B models, respectively.  Hierarchical merging with incorporated context  overwhelmingly outperforms comparison systems without supporting evidence (see first block in the tables). For each metric (and dataset) the best performing model is  a variant of hierarchical merging with context (the only exception is BERTScore on  SuperSummary for Cite-HMerge using the 8B model). 

\paragraph{Extract-Support is the best model.}
Among the proposed approaches, Extract-Support demonstrates superior performance, achieving the highest average score across all metrics. Cite emerges as the least effective context selection method, showing superior performance only in AlignScore with minor differences to second best results. 
When examining the impact of model size, scaling to 70B parameters yields moderate improvements on reference-based metrics, although gains on input-based metrics are less substantial. Additionally, our analysis indicates that models consistently produce lower scores on SuperSummary compared to Multi-LexSum, suggesting that book summarization poses greater challenges than legal document summarization.

\paragraph{Replace models show big improvements according to input-based metrics.}
Examining input-based metrics more closely, both SummaC and AlignScore show significant improvements when input passages are incorporated via Replace. Specifically, for the 8B model, while Support achieves similar results to baselines, Replace consistently demonstrates superior performance, with Retrieve-Replace and Cite-Replace achieving approximately 10-points improvements in AlignScore on SuperSummary compared to other methods. The same patterns are again observed in the 70B model results, where Retrieve-Replace excels in SummaC scores and Cite-Replace leads in AlignScore metrics. These results align with our expectations, given that both metrics evaluate performance against input texts, and Replace explicitly leverages input context in summary generation.

\begin{table}[t]
  \centering
  \renewcommand{\arraystretch}{1.5}
  \fontsize{9.5}{9}\selectfont
  \begin{tabular}{@{}l@{}c@{~}c@{~}c@{~}c@{~}c@{}}
    \hline
    \multicolumn{6}{c}{\large{\textbf{Multi-LexSum}}} \\
Models   &  {ROUGE} & BScore & \textsc{SummaC} & AScore & \textsc{PRisma} \\
    \hline
    Zero-shot & 24.6 & 60.6 & 40.9 & 79.3 & 42.7 \\
    HMerge & 24.4 & 61.7 & 41.1 & 72.9 & 45.5 \\
    Cite-HMerge & 23.2 & 62.4 & 43.3 & 82.0 & 43.0 \\
    \hline
    Retrieve-R & 23.4 & 61.5 & \textbf{47.0} & \textbf{84.6} & 41.8 \\
    Retrieve-S & \uline{24.9} & \uline{62.6} & 41.2 & 79.4 & \uline{46.1} \\
    \hline
    Extract-R & 24.1 & 61.6 & 43.2 & 79.4 & 41.9 \\
    Extract-S & \textbf{25.8} & \textbf{63.3} & 40.6 & 79.4 & \textbf{47.5} \\
    \hline
    Cite-R & 20.4 & 59.2 & \uline{46.8} & \uline{82.3} & 36.3 \\
    Cite-S & 23.0 & 61.5 & 42.7 & 81.0 & 43.0 \\
    \hline
    \multicolumn{6}{c}{} \\ \hline
    \multicolumn{6}{c}{\large{\textbf{SuperSummary}}}\\
    & {ROUGE} & BScore & \textsc{SummaC} & AScore & \textsc{PRisma} \\
    \hline
    Zero-shot & 18.1 & 57.8 & 38.8 & 60.7 & 33.2 \\
    HM & \uline{20.3} & 63.3 & 38.7 & 55.8 & 37.8 \\
    Cite-HMerge & 19.8 & \textbf{66.0} & 38.1 & 59.1 & 35.1 \\
    \hline
    Retrieve-R & 16.5 & 60.0 & \textbf{47.6} & \uline{70.2} & 23.3 \\
    Retrieve-S & 19.9 & \uline{65.4} & 38.6 & 56.4 & \uline{38.4} \\
    \hline
    Extract-R & 16.8 & 58.1 & \uline{41.6} & 61.1 & 22.3 \\
    Extract-S & \textbf{21.2} & 63.7 & 38.3 & 57.2 & \textbf{39.2} \\
    \hline
    Cite-R & 15.6 & 58.1 & 40.4 & \textbf{71.8} & 20.3 \\
   Cite-S & 19.3 & 62.3 & 37.9 & 55.1 & 34.6 \\
    \hline
  \end{tabular}
  \caption{Results with Llama-3.1-8B-Instruct model. We report the
    geometric mean of ROUGE-1, \mbox{ROUGE-2}, and ROUGEL-L (ROUGE). BScore
    and AScore are shorthands for BERTScore and AlignScore,
    respectively. Labels~-R and~-S, refer to Replace and Support
    context incorporation methods. Best results for each metric are
    highlighted in \textbf{bold}, whereas the second best results for
    each metric are \uline{underlined}.}
  \label{tab:8b}
\end{table}

\begin{table}[t]
  \centering
  \renewcommand{\arraystretch}{1.5}
  \fontsize{9.5}{9}\selectfont
  \begin{tabular}{@{}l@{~}c@{~}c@{~}c@{~}c@{~}c@{}}
    \hline
    \multicolumn{6}{c}{\large{\textbf{Multi-LexSum}}}\\
    Models & {ROUGE} & BScore & \textsc{SummaC} & AScore & \textsc{PRisma} \\
    \hline
    Zero-shot & 23.6 & 60.7 & 43.5 & 77.6 & 41.5 \\
    HMerge & \uline{26.7} & \uline{64.3} & 43.4 & 76.3 & 48.2 \\
    Cite-HMerge & 25.6 & 64.0 & 43.1 & 82.7 & 47.3 \\
    \hline
    Retrieve-R & 24.7 & 62.1 & \uline{47.9} & 80.3 & 43.8 \\
    Retrieve-S & 26.6 & \textbf{66.1} & 44.5 & 78.8 & \textbf{49.8} \\
    \hline
    Extract-R & 24.8 & 60.4 & 43.4 & 79.9 & 42.7 \\
    Extract-S & \textbf{27.6} & 64.1 & 43.2 & 79.0 & \uline{49.7} \\
    \hline
    Cite-R & 22.5 & 61.9 & \textbf{51.6} & \textbf{85.8} & 40.6 \\
    Cite-S & 25.4 & 62.5 & 44.4 & \uline{83.9} & 46.8 \\
    \hline
\multicolumn{6}{c}{} \\ \hline
    \multicolumn{6}{c}{\large{\textbf{SuperSummary}}} \\
    Models & {ROUGE} & BScore & \textsc{SummaC} & AScore & \textsc{PRisma} \\
    \hline
    Zero-shot & 18.9 & 58.1 & 38.4 & 57.9 & 35.2 \\
    HMerge & 21.8 & 65.4 & 39.1 & 57.8 & 42.2 \\
    Cite-HMerge & 20.9 & \uline{66.7} & 38.4 & 59.2 & 38.8 \\
    \hline
    Retrieve-R & 16.9 & 58.9 & \textbf{44.4} & \uline{66.7} & 25.4 \\
    Retrieve-S & \uline{22.4} & 65.5 & 38.7 & 59.3 & \uline{44.3} \\
    \hline
    Extract-R & 17.2 & 59.0 & 41.6 & 59.7 & 23.5 \\
    Extract-S & \textbf{22.9} & \textbf{67.2} & 37.6 & 58.5 & \textbf{45.6} \\
    \hline
    Cite-R & 15.5 & 59.9 & \uline{43.6} & \textbf{78.4} & 21.8 \\
    Cite-S & 20.6 & 65.3 & 37.8 & 57.0 & 40.8 \\
    \hline
  \end{tabular}
  \caption{Results for the Llama-3.1-70B-Instruct model. We report the
    geometric mean of ROUGE-1, \mbox{ROUGE-2}, and ROUGEL-L
    (ROUGE). BScore and AScore are shorthands for BERTScore and
    AlignScore, respectively. Labels -R and -S, refer to Replace and
    Support context incorporation methods Best results for each metric
    are highlighted in \textbf{bold}, while the second best results
    for each metric are \uline{underlined}.}
  \label{tab:70b}
\end{table}

\paragraph{Support models are better summarizers according to reference-based metrics.}
Reference-based metrics reveal patterns that stand in sharp contrast to those observed with input-based metrics. In both 8B and 70B models, Extract-Support and Retrieve-Support consistently outperform baselines. However, Cite-Support shows comparable or slightly worse performance, suggesting that while input contexts can enhance summary faithfulness, their effectiveness depends heavily on the method's ability to identify and utilize relevant segments. Compared to Support methods, Replace methods show significantly inferior performance across both model scales, with particularly pronounced differences on SuperSummary, where Replace methods lag behind Support by approximately 20 points across all three context selection methods in 70B models. This performance gap is further corroborated by both ROUGE and BERTScore metrics. In summary, results with reference-based metrics strongly suggest that methods incorporating abstractive summaries during hierarchical merging (as in Support and baseline approaches) are better at maintaining comprehensive coverage compared to methods that rely solely on input contexts for final summary generation (Replace approaches), which may become overly specific and miss broader aspects of the source material.


\paragraph{Adding more contexts yields better summaries for Replace.} Our results in Tables~\ref{tab:8b} and~\ref{tab:70b} suggest that incorporating contexts through Replace degrades performance across reference-based metrics. Manual inspection of Replace summaries further showed that they tend to emphasize different key events compared to Support or baseline methods, often prioritizing peripheral details such as casual dialogue or scene descriptions. Despite having comparable lengths, final-stage inputs from Support and baseline methods have higher information density compared to Replace.  Specifically, Replace contexts typically elaborate on a single key event extensively, while abstractive summaries convey the same event more concisely, allowing space for additional key events.


Building on these observations, we investigate whether Replace summaries can be improved by increasing the maximum merging context length, thereby incorporating more information while merging the same number of summaries. Specifically, we extend the maximum merging context length from 8K to 16K and 32K tokens. We maintain the same retrieval units but select more passages that fit within this increased context length window, thereby generating next-level summaries with greater information density. We experiment with context extension on the 8B model using both datasets, employing Retrieve-Replace as the hierarchical merging method. The results, shown in Table~\ref{tab:ret_expand}, indicate that using 16K and 32K tokens achieves higher \textsc{PRisma} scores on both datasets, with more substantial gains on SuperSummary compared to Multi-LexSum. However, even with the increased threshold of 32K tokens,  performance still falls short of the results achieved by Retrieve-Support, demonstrating that the use of abstractive summaries is critical  for achieving better coverage.

\begin{table}[t]
  \centering
  \renewcommand{\arraystretch}{1.5}
  \fontsize{10}{10}\selectfont
  \small
  \begin{tabular}{clclc}
    \hline
   \multicolumn{1}{c}{Context Length} & \multicolumn{2}{c}{\textbf{Multi-LexSum}} & \multicolumn{2}{c}{\textbf{SuperSummary}} \\
    \hline
    8K & 41.8 &--- & 23.3  &--- \\
    16K & 43.7 &\textcolor{red}{$+$1.9} & 26.5 \hspace{0.1cm} &\textcolor{red}{$+$3.2} \\
    32K & 44.1 & \textcolor{red}{$+$2.3} & 27.9 & \textcolor{red}{$+$4.6} \\
    \hline
  \end{tabular}

  \caption{\textsc{PRisma} results for Retrieve-Replace with different maximum merging context lengths Llama-3.1-8B-Instruct. Gains compared to 8K context length are shown in red. }
  \label{tab:ret_expand}
\end{table}

\paragraph{Manual analysis confirms Extract-Support generates the most
  faithful summaries.} Our results suggest that input- and
reference-based metrics tend to favor different methodological
approaches. To evaluate metric reliability, we analyzed book summaries
generated by the 70B model. Specifically, we decomposed output
summaries for the book \emph{Things we Hide from the Light} (by Lucy
Score) into atomic claims and verified whether they are supported by
the source. We followed a simplified version of the annotation
guidelines proposed in \citet{fables-2024-kim-et-al}. Claims were
classified as `Correct' if fully supported by the book, `Incorrect' if
parts of the claim contradict the source, or `Not Present' if parts of
the claim are absent from the source material.

Table~\ref{tab:qualtitative_analysis} presents the results of our
analysis for Extract-Support, Extract-Replace, and the three
comparison  baselines (Cite-HMerge, Zero-Shot, and HMerge). The
summaries and claim annotations can be found in
Appendix~\ref{sec:example_summaries} and~\ref{sec:facts_verify},
respectively. We observe that Extract-Support summaries contain the
highest proportion of Correct claims (10\% higher  compared to
baseline methods). When comparing  claims from Extract-Support and the
original HMerge, we find that while the summaries maintain overall
similar structure, there are notable differences in the selection of key events. These differences can be attributed to the incorporation of supporting evidence, which may influence the prioritization of events mentioned in such contexts.

In contrast, Extract-Replace performs below baseline levels, by more than~10\%. This poor performance may be explained by inherent challenges in extractive summarization, where the merging of input contexts can introduce coherence issues that potentially confuse LLMs, leading to unfaithful summary generation \cite{zhang-etal-2023-extractive}. The relative performance of these methods (Extract-Support being most factual and Extract-Replace being least factual) aligns with automatic results using reference-based metrics such as \textsc{PRisma} (see last column in Tables~\ref{tab:8b} and~\ref{tab:70b}). Taken together, human-based and automatic evaluation results suggest that comprehensive source coverage contributes to enhanced faithfulness in the final summary.

\begin{table}[t]
  \centering
  \renewcommand{\arraystretch}{1.5}
  \fontsize{9}{9}\selectfont
    \begin{tabular}{lcc|c}
      \hline
      \multicolumn{1}{c}{Models} & Correct & Incorrect & Not Present \\
      \hline
      Extract-Support & 72.7\% & 18.2\% & \hspace{.14cm}9.1\% \\
      Cite-HMerge & 60.0\% & 20.0\% & 20.0\% \\
      Zero-shot & 60.0\% & 20.0\% & 20.0\% \\
      HMerge & 59.1\% & 27.3\% & 13.6\% \\
      Extract-Replace & 48.8\% & 23.3\% & 27.9\% \\
      \hline
    \end{tabular}
  \caption{Proportion of Correct, Incorrect, and Not Present claims for summaries generated by Extract-Support, Extract-Replace, and comparison baselines using Llama-3.1-70B-Instruct (GPTQ-INT4). Models  summarize the book \emph{Things we Hide from the Light} (by Lucy Score) from the SuperSummary dataset.}
   
  \label{tab:qualtitative_analysis}
\end{table}

\section{Conclusion}

In this paper, we examined how input contexts can be integrated into hierarchical merging so as  to produce more faithful summaries for very long source documents (>100K tokens). 

We proposed different approaches to   contextual augmentation ranging
from \emph{replacing} intermediate summaries with
relevant input context, to \emph{refining} them while
using the context as supporting evidence, and
implicitly \emph{aligning} summary sentences (via citations) to 
input passages. Our findings indicate that contextual augmentation leads to more factual summaries. However, input contexts alone can distract from important information, emphasizing peripheral details, and thus work best to  support  rather than substitute abstractive summaries that capture a broader perspective of the source content. 

In the future, we would like to learn when to perform context augmentation based on the quality of the intermediate summary (i.e., calling  context augmentation as a tool) and to explore trade-offs between context length and summary quality. 

\section{Limitations}

\paragraph{Large-scale Human Evaluation} We verify automatic evaluation results with human annotations
performed on model summaries corresponding to one book. Human
evaluation results could be enhanced by performing experiments on
multiple books, using a bigger annotator pool. There are
well-documented challenges associated with human-based evaluations of
summaries based on very long input documents
\cite{chang2024booookscore,krishna-etal-2023-longeval,xu-etal-2023-critical,fables-2024-kim-et-al}. In
our case, hiring multiple annotators to read and understand books (or
legal documents) would have been prohibitively expensive and
time-consuming. \citet{fables-2024-kim-et-al} report annotation costs
in the range of~\$200 to \$250 per book, requiring 10 hours of work
(excluding the time it takes to read the book).


\paragraph{Efficient Run-time for Support} Support summaries are
generated using input contexts as supporting evidence. This results in
longer inputs for LLMs during the merging phase compared to approaches
that do not include supporting evidence. Future research could explore
more efficient ways of incorporating contextual information into
the summarization process, by selectively including 
context, e.g.,~based on positional information or knowledge about the
structure of the input.

\paragraph{Limited Dataset Domain} This work focuses on the specific task of full-text summarization for very long documents (>100K tokens). This constraint significantly limits the available datasets, restricting both the domains and samples we can use to test our proposed pipeline. Although we identified other potentially suitable datasets, such as LFOSum \cite{nayeem2024lfosumsummarizinglongformopinions}, which focuses on opinion summarization with an average input length of 207K tokens, these were not open-sourced. Overall, there is a pressing need to curate more high-quality summarization datasets containing inputs exceeding 100K tokens across diverse domains.

\paragraph{The Role of Context} Our results so far have shown that
incorporating context generally improves the faithfulness of the
generated summaries. More fine-grained analysis is required to
establish how contexts improve the quality of the intermediate
summaries as well as the end result. For example, it would be interesting to study how to dynamically select contexts to increase efficiency and avoid unnecessary distraction from using contexts at merges that don't need them.

\section*{Acknowledgments}

We thank the anonymous reviewers for their constructive feedback. We gratefully acknowledge the support of the UK Engineering and Physical Sciences
Research Council (grant EP/W002876/1).
\bibliography{custom}

\begin{thebibliography}{51}
\providecommand{\natexlab}[1]{#1}

\bibitem[{Andryushchenko et~al.(2024)Andryushchenko, Ivanov, Makharev,
  Tukhtina, and Valeev}]{andryushchenko2024leveraginglargelanguagemodels}
Georgy Andryushchenko, Vladimir Ivanov, Vladimir Makharev, Elizaveta Tukhtina,
  and Aidar Valeev. 2024.
\newblock \href {https://arxiv.org/abs/2411.03012} {Leveraging large language
  models in code question answering: Baselines and issues}.
\newblock \emph{Preprint}, arXiv:2411.03012.

\bibitem[{Asai et~al.(2023)Asai, Wu, Wang, Sil, and
  Hajishirzi}]{asai2023selfraglearningretrievegenerate}
Akari Asai, Zeqiu Wu, Yizhong Wang, Avirup Sil, and Hannaneh Hajishirzi. 2023.
\newblock \href {https://arxiv.org/abs/2310.11511} {Self-rag: Learning to
  retrieve, generate, and critique through self-reflection}.
\newblock \emph{Preprint}, arXiv:2310.11511.

\bibitem[{Asai et~al.(2024)Asai, Zhong, Chen, Koh, Zettlemoyer, Hajishirzi, and
  tau Yih}]{asai2024reliableadaptableattributablelanguage}
Akari Asai, Zexuan Zhong, Danqi Chen, Pang~Wei Koh, Luke Zettlemoyer, Hannaneh
  Hajishirzi, and Wen tau Yih. 2024.
\newblock \href {https://arxiv.org/abs/2403.03187} {Reliable, adaptable, and
  attributable language models with retrieval}.
\newblock \emph{Preprint}, arXiv:2403.03187.

\bibitem[{Beltagy et~al.(2020)Beltagy, Peters, and
  Cohan}]{beltagy2020longformerlongdocumenttransformer}
Iz~Beltagy, Matthew~E. Peters, and Arman Cohan. 2020.
\newblock \href {https://arxiv.org/abs/2004.05150} {Longformer: The
  long-document transformer}.
\newblock \emph{Preprint}, arXiv:2004.05150.

\bibitem[{Bertsch et~al.(2023)Bertsch, Alon, Neubig, and
  Gormley}]{bertsch2023unlimiformer}
Amanda Bertsch, Uri Alon, Graham Neubig, and Matthew~R Gormley. 2023.
\newblock Unlimiformer: Long-range transformers with unlimited length input.
\newblock \emph{arXiv preprint arXiv:2305.01625}.

\bibitem[{Bian et~al.(2023)Bian, Huang, Zhou, and
  Zhu}]{bian2023gosumextractivesummarizationlong}
Junyi Bian, Xiaodi Huang, Hong Zhou, and Shanfeng Zhu. 2023.
\newblock \href {https://arxiv.org/abs/2211.10247} {Gosum: Extractive
  summarization of long documents by reinforcement learning and graph organized
  discourse state}.
\newblock \emph{Preprint}, arXiv:2211.10247.

\bibitem[{Bohnet et~al.(2022)Bohnet, Tran, Verga, Aharoni, Andor, Soares,
  Ciaramita, Eisenstein, Ganchev, Herzig, Hui, Kwiatkowski, Ma, Ni, Schuster,
  Saralegui, Cohen, Collins, Das, Metzler, Petrov, and
  Webster}]{Bohnet:Ea:2022}
Bernd Bohnet, Vinh Tran, Pat Verga, Roee Aharoni, Daniel Andor, Livio~Baldini
  Soares, Massimiliano Ciaramita, Jacob Eisenstein, Kuzman Ganchev, Jonathan
  Herzig, Kai Hui, Tom Kwiatkowski, Ji~Ma, Jianmo Ni, Tal Schuster,
  Lierni~Sestorain Saralegui, William~Weston Cohen, Michael Collins, Dipanjan
  Das, Don Metzler, Slav Petrov, and Kellie Webster. 2022.
\newblock \href {https://arxiv.org/abs/2212.08037} {Attributed question
  answering: Evaluation and modeling for attributed large language models}.

\bibitem[{Chang et~al.(2024)Chang, Lo, Goyal, and Iyyer}]{chang2024booookscore}
Yapei Chang, Kyle Lo, Tanya Goyal, and Mohit Iyyer. 2024.
\newblock \href {https://arxiv.org/pdf/2310.00785.pdf} {Booookscore: A
  systematic exploration of book-length summarization in the era of {LLM}s}.
\newblock In \emph{The Twelfth International Conference on Learning
  Representations}.

\bibitem[{Chen et~al.(2023)Chen, Wong, Chen, and
  Tian}]{chen2023extendingcontextwindowlarge}
Shouyuan Chen, Sherman Wong, Liangjian Chen, and Yuandong Tian. 2023.
\newblock \href {https://arxiv.org/abs/2306.15595} {Extending context window of
  large language models via positional interpolation}.
\newblock \emph{Preprint}, arXiv:2306.15595.

\bibitem[{Chen et~al.(2024)Chen, Qian, Tang, Lai, Liu, Han, and
  Jia}]{chen2024longloraefficientfinetuninglongcontext}
Yukang Chen, Shengju Qian, Haotian Tang, Xin Lai, Zhijian Liu, Song Han, and
  Jiaya Jia. 2024.
\newblock \href {https://arxiv.org/abs/2309.12307} {Longlora: Efficient
  fine-tuning of long-context large language models}.
\newblock \emph{Preprint}, arXiv:2309.12307.

\bibitem[{Child et~al.(2019)Child, Gray, Radford, and
  Sutskever}]{child2019generatinglongsequencessparse}
Rewon Child, Scott Gray, Alec Radford, and Ilya Sutskever. 2019.
\newblock \href {https://arxiv.org/abs/1904.10509} {Generating long sequences
  with sparse transformers}.
\newblock \emph{Preprint}, arXiv:1904.10509.

\bibitem[{Cohan et~al.(2018)Cohan, Dernoncourt, Kim, Bui, Kim, Chang, and
  Goharian}]{cohan-etal-2018-discourse}
Arman Cohan, Franck Dernoncourt, Doo~Soon Kim, Trung Bui, Seokhwan Kim, Walter
  Chang, and Nazli Goharian. 2018.
\newblock \href {https://doi.org/10.18653/v1/N18-2097} {A discourse-aware
  attention model for abstractive summarization of long documents}.
\newblock In \emph{Proceedings of the 2018 Conference of the North {A}merican
  Chapter of the Association for Computational Linguistics: Human Language
  Technologies, Volume 2 (Short Papers)}, pages 615--621, New Orleans,
  Louisiana. Association for Computational Linguistics.

\bibitem[{Edge et~al.(2024)Edge, Trinh, Cheng, Bradley, Chao, Mody, Truitt, and
  Larson}]{edge2024localglobalgraphrag}
Darren Edge, Ha~Trinh, Newman Cheng, Joshua Bradley, Alex Chao, Apurva Mody,
  Steven Truitt, and Jonathan Larson. 2024.
\newblock \href {https://arxiv.org/abs/2404.16130} {From local to global: A
  graph rag approach to query-focused summarization}.
\newblock \emph{Preprint}, arXiv:2404.16130.

\bibitem[{Fierro et~al.(2024)Fierro, Amplayo, Huot, Cao, Maynez, Narayan, and
  Lapata}]{fierro2024learningplangeneratetext}
Constanza Fierro, Reinald~Kim Amplayo, Fantine Huot, Nicola~De Cao, Joshua
  Maynez, Shashi Narayan, and Mirella Lapata. 2024.
\newblock \href {https://arxiv.org/abs/2404.03381} {Learning to plan and
  generate text with citations}.
\newblock \emph{Preprint}, arXiv:2404.03381.

\bibitem[{Frantar et~al.(2023)Frantar, Ashkboos, Hoefler, and
  Alistarh}]{frantar2023gptqaccurateposttrainingquantization}
Elias Frantar, Saleh Ashkboos, Torsten Hoefler, and Dan Alistarh. 2023.
\newblock \href {https://arxiv.org/abs/2210.17323} {Gptq: Accurate
  post-training quantization for generative pre-trained transformers}.
\newblock \emph{Preprint}, arXiv:2210.17323.

\bibitem[{Gao et~al.(2023)Gao, Yen, Yu, and Chen}]{gao-etal-2023-enabling}
Tianyu Gao, Howard Yen, Jiatong Yu, and Danqi Chen. 2023.
\newblock \href {https://doi.org/10.18653/v1/2023.emnlp-main.398} {Enabling
  large language models to generate text with citations}.
\newblock In \emph{Proceedings of the 2023 Conference on Empirical Methods in
  Natural Language Processing}, pages 6465--6488, Singapore. Association for
  Computational Linguistics.

\bibitem[{Gu et~al.(2022)Gu, Ash, and Hahnloser}]{gu-etal-2022-memsum}
Nianlong Gu, Elliott Ash, and Richard Hahnloser. 2022.
\newblock \href {https://aclanthology.org/2022.acl-long.450} {{M}em{S}um:
  Extractive summarization of long documents using multi-step episodic {M}arkov
  decision processes}.
\newblock In \emph{Proceedings of the 60th Annual Meeting of the Association
  for Computational Linguistics (Volume 1: Long Papers)}, pages 6507--6522,
  Dublin, Ireland. Association for Computational Linguistics.

\bibitem[{He et~al.(2021)He, Liu, Gao, and
  Chen}]{he2021debertadecodingenhancedbertdisentangled}
Pengcheng He, Xiaodong Liu, Jianfeng Gao, and Weizhu Chen. 2021.
\newblock \href {https://arxiv.org/abs/2006.03654} {Deberta: Decoding-enhanced
  bert with disentangled attention}.
\newblock \emph{Preprint}, arXiv:2006.03654.

\bibitem[{Hemamou and Debiane(2024)}]{Hemamou2024ScalingUS}
L'eo Hemamou and Mehdi Debiane. 2024.
\newblock \href {https://api.semanticscholar.org/CorpusID:271974468} {Scaling
  up summarization: Leveraging large language models for long text extractive
  summarization}.
\newblock \emph{ArXiv}, abs/2408.15801.

\bibitem[{Huang et~al.(2021)Huang, Cao, Parulian, Ji, and
  Wang}]{huang-etal-2021-govreport}
Luyang Huang, Shuyang Cao, Nikolaus Parulian, Heng Ji, and Lu~Wang. 2021.
\newblock \href {https://doi.org/10.18653/v1/2021.naacl-main.112} {Efficient
  attentions for long document summarization}.
\newblock In \emph{Proceedings of the 2021 Conference of the North American
  Chapter of the Association for Computational Linguistics: Human Language
  Technologies}, pages 1419--1436, Online. Association for Computational
  Linguistics.

\bibitem[{Izacard and Grave(2021)}]{izacard-grave-2021-leveraging}
Gautier Izacard and Edouard Grave. 2021.
\newblock \href {https://doi.org/10.18653/v1/2021.eacl-main.74} {Leveraging
  passage retrieval with generative models for open domain question answering}.
\newblock In \emph{Proceedings of the 16th Conference of the European Chapter
  of the Association for Computational Linguistics: Main Volume}, pages
  874--880, Online. Association for Computational Linguistics.

\bibitem[{Katharopoulos et~al.(2020)Katharopoulos, Vyas, Pappas, and
  Fleuret}]{katharopoulos:ea:2020}
Angelos Katharopoulos, Apoorv Vyas, Nikolaos Pappas, and Fran\c{c}ois Fleuret.
  2020.
\newblock Transformers are rnns: fast autoregressive transformers with linear
  attention.
\newblock In \emph{Proceedings of the 37th International Conference on Machine
  Learning}, ICML'20. JMLR.org.

\bibitem[{Kim et~al.(2024)Kim, Chang, Karpinska, Garimella, Manjunatha, Lo,
  Goyal, and Iyyer}]{fables-2024-kim-et-al}
Yekyung Kim, Yapei Chang, Marzena Karpinska, Aparna Garimella, Varun
  Manjunatha, Kyle Lo, Tanya Goyal, and Mohit Iyyer. 2024.
\newblock \href {https://arxiv.org/abs/2404.01261} {Fables: Evaluating
  faithfulness and content selection in book-length summarization}.
\newblock \emph{Preprint}, arXiv:2404.01261.

\bibitem[{Krishna et~al.(2023)Krishna, Bransom, Kuehl, Iyyer, Dasigi, Cohan,
  and Lo}]{krishna-etal-2023-longeval}
Kalpesh Krishna, Erin Bransom, Bailey Kuehl, Mohit Iyyer, Pradeep Dasigi, Arman
  Cohan, and Kyle Lo. 2023.
\newblock \href {https://doi.org/10.18653/v1/2023.eacl-main.121} {{L}ong{E}val:
  Guidelines for human evaluation of faithfulness in long-form summarization}.
\newblock In \emph{Proceedings of the 17th Conference of the European Chapter
  of the Association for Computational Linguistics}, pages 1650--1669,
  Dubrovnik, Croatia. Association for Computational Linguistics.

\bibitem[{Kryscinski et~al.(2022)Kryscinski, Rajani, Agarwal, Xiong, and
  Radev}]{kryscinski-etal-2022-booksum}
Wojciech Kryscinski, Nazneen Rajani, Divyansh Agarwal, Caiming Xiong, and
  Dragomir Radev. 2022.
\newblock \href {https://doi.org/10.18653/v1/2022.findings-emnlp.488}
  {{BOOKSUM}: A collection of datasets for long-form narrative summarization}.
\newblock In \emph{Findings of the Association for Computational Linguistics:
  EMNLP 2022}, pages 6536--6558, Abu Dhabi, United Arab Emirates. Association
  for Computational Linguistics.

\bibitem[{Laban et~al.(2021)Laban, Schnabel, Bennett, and
  Hearst}]{laban2021summacrevisitingnlibasedmodels}
Philippe Laban, Tobias Schnabel, Paul~N. Bennett, and Marti~A. Hearst. 2021.
\newblock \href {https://arxiv.org/abs/2111.09525} {Summac: Re-visiting
  nli-based models for inconsistency detection in summarization}.
\newblock \emph{Preprint}, arXiv:2111.09525.

\bibitem[{Lewis et~al.(2020)Lewis, Perez, Piktus, Petroni, Karpukhin, Goyal,
  K\"{u}ttler, Lewis, Yih, Rockt\"{a}schel, Riedel, and Kiela}]{lewis2020rag}
Patrick Lewis, Ethan Perez, Aleksandra Piktus, Fabio Petroni, Vladimir
  Karpukhin, Naman Goyal, Heinrich K\"{u}ttler, Mike Lewis, Wen-tau Yih, Tim
  Rockt\"{a}schel, Sebastian Riedel, and Douwe Kiela. 2020.
\newblock \href
  {https://proceedings.neurips.cc/paper_files/paper/2020/file/6b493230205f780e1bc26945df7481e5-Paper.pdf}
  {Retrieval-augmented generation for knowledge-intensive nlp tasks}.
\newblock In \emph{Advances in Neural Information Processing Systems},
  volume~33, pages 9459--9474. Curran Associates, Inc.

\bibitem[{Li et~al.(2024)Li, Zhang, Do, Yue, and
  Chen}]{li2024longcontextllmsstrugglelong}
Tianle Li, Ge~Zhang, Quy~Duc Do, Xiang Yue, and Wenhu Chen. 2024.
\newblock \href {https://arxiv.org/abs/2404.02060} {Long-context llms struggle
  with long in-context learning}.
\newblock \emph{Preprint}, arXiv:2404.02060.

\bibitem[{Lin(2004)}]{lin-2004-rouge}
Chin-Yew Lin. 2004.
\newblock \href {https://aclanthology.org/W04-1013} {{ROUGE}: A package for
  automatic evaluation of summaries}.
\newblock In \emph{Text Summarization Branches Out}, pages 74--81, Barcelona,
  Spain. Association for Computational Linguistics.

\bibitem[{Liu and Lapata(2019)}]{liu-lapata-2019-text}
Yang Liu and Mirella Lapata. 2019.
\newblock \href {https://doi.org/10.18653/v1/D19-1387} {Text summarization with
  pretrained encoders}.
\newblock In \emph{Proceedings of the 2019 Conference on Empirical Methods in
  Natural Language Processing and the 9th International Joint Conference on
  Natural Language Processing (EMNLP-IJCNLP)}, pages 3730--3740, Hong Kong,
  China. Association for Computational Linguistics.

\bibitem[{Lu et~al.(2023)Lu, Larcher, and Tran}]{Lu2023HybridLD}
Guang Lu, Sylvia~B. Larcher, and Tu-Anh Tran. 2023.
\newblock \href {https://api.semanticscholar.org/CorpusID:259064158} {Hybrid
  long document summarization using c2f-far and chatgpt: A practical study}.
\newblock \emph{ArXiv}, abs/2306.01169.

\bibitem[{Mahon and Lapata(2024)}]{mahon-lapata-2024-modular}
Louis Mahon and Mirella Lapata. 2024.
\newblock \href {https://doi.org/10.18653/v1/2024.acl-long.450} {A modular
  approach for multimodal summarization of {TV} shows}.
\newblock In \emph{Proceedings of the 62nd Annual Meeting of the Association
  for Computational Linguistics (Volume 1: Long Papers)}, pages 8272--8291,
  Bangkok, Thailand. Association for Computational Linguistics.

\bibitem[{Menick et~al.(2022)Menick, Trebacz, Mikulik, Aslanides, Song,
  Chadwick, Glaese, Young, Campbell-Gillingham, Irving, and
  McAleese}]{menick2022teachinglanguagemodelssupport}
Jacob Menick, Maja Trebacz, Vladimir Mikulik, John Aslanides, Francis Song,
  Martin Chadwick, Mia Glaese, Susannah Young, Lucy Campbell-Gillingham,
  Geoffrey Irving, and Nat McAleese. 2022.
\newblock \href {https://arxiv.org/abs/2203.11147} {Teaching language models to
  support answers with verified quotes}.
\newblock \emph{Preprint}, arXiv:2203.11147.

\bibitem[{Min et~al.(2023)Min, Krishna, Lyu, Lewis, Yih, Koh, Iyyer,
  Zettlemoyer, and Hajishirzi}]{min-etal-2023-factscore}
Sewon Min, Kalpesh Krishna, Xinxi Lyu, Mike Lewis, Wen-tau Yih, Pang Koh, Mohit
  Iyyer, Luke Zettlemoyer, and Hannaneh Hajishirzi. 2023.
\newblock \href {https://doi.org/10.18653/v1/2023.emnlp-main.741}
  {{FA}ct{S}core: Fine-grained atomic evaluation of factual precision in long
  form text generation}.
\newblock In \emph{Proceedings of the 2023 Conference on Empirical Methods in
  Natural Language Processing}, pages 12076--12100, Singapore. Association for
  Computational Linguistics.

\bibitem[{Nakano et~al.(2022)Nakano, Hilton, Balaji, Wu, Ouyang, Kim, Hesse,
  Jain, Kosaraju, Saunders, Jiang, Cobbe, Eloundou, Krueger, Button, Knight,
  Chess, and Schulman}]{nakano2022webgptbrowserassistedquestionansweringhuman}
Reiichiro Nakano, Jacob Hilton, Suchir Balaji, Jeff Wu, Long Ouyang, Christina
  Kim, Christopher Hesse, Shantanu Jain, Vineet Kosaraju, William Saunders,
  Xu~Jiang, Karl Cobbe, Tyna Eloundou, Gretchen Krueger, Kevin Button, Matthew
  Knight, Benjamin Chess, and John Schulman. 2022.
\newblock \href {https://arxiv.org/abs/2112.09332} {Webgpt: Browser-assisted
  question-answering with human feedback}.
\newblock \emph{Preprint}, arXiv:2112.09332.

\bibitem[{Narayan et~al.(2018)Narayan, Cohen, and
  Lapata}]{narayan-etal-2018-ranking}
Shashi Narayan, Shay~B. Cohen, and Mirella Lapata. 2018.
\newblock \href {https://doi.org/10.18653/v1/N18-1158} {Ranking sentences for
  extractive summarization with reinforcement learning}.
\newblock In \emph{Proceedings of the 2018 Conference of the North {A}merican
  Chapter of the Association for Computational Linguistics: Human Language
  Technologies, Volume 1 (Long Papers)}, pages 1747--1759, New Orleans,
  Louisiana. Association for Computational Linguistics.

\bibitem[{Nayeem and
  Rafiei(2024)}]{nayeem2024lfosumsummarizinglongformopinions}
Mir~Tafseer Nayeem and Davood Rafiei. 2024.
\newblock \href {https://arxiv.org/abs/2410.13037} {Lfosum: Summarizing
  long-form opinions with large language models}.
\newblock \emph{Preprint}, arXiv:2410.13037.

\bibitem[{Peng et~al.(2023)Peng, Quesnelle, Fan, and
  Shippole}]{peng2023yarnefficientcontextwindow}
Bowen Peng, Jeffrey Quesnelle, Honglu Fan, and Enrico Shippole. 2023.
\newblock \href {https://arxiv.org/abs/2309.00071} {Yarn: Efficient context
  window extension of large language models}.
\newblock \emph{Preprint}, arXiv:2309.00071.

\bibitem[{Shaham et~al.(2023)Shaham, Ivgi, Efrat, Berant, and
  Levy}]{shaham-etal-2023-zeroscrolls}
Uri Shaham, Maor Ivgi, Avia Efrat, Jonathan Berant, and Omer Levy. 2023.
\newblock \href {https://doi.org/10.18653/v1/2023.findings-emnlp.536}
  {{Z}ero{SCROLLS}: A zero-shot benchmark for long text understanding}.
\newblock In \emph{Findings of the Association for Computational Linguistics:
  EMNLP 2023}, pages 7977--7989, Singapore. Association for Computational
  Linguistics.

\bibitem[{Shen et~al.(2022)Shen, Lo, Yu, Dahlberg, Schlanger, and
  Downey}]{Shen2022MultiLexSum}
Zejiang Shen, Kyle Lo, Lauren Yu, Nathan Dahlberg, Margo Schlanger, and Doug
  Downey. 2022.
\newblock \href {https://doi.org/10.48550/arXiv.2206.10883} {Multi-lexsum:
  Real-world summaries of civil rights lawsuits at multiple granularities}.
\newblock \emph{CoRR}, abs/2206.10883.

\bibitem[{Shi et~al.(2023)Shi, Min, Yasunaga, Seo, James, Lewis, Zettlemoyer,
  and tau Yih}]{shi2023replugretrievalaugmentedblackboxlanguage}
Weijia Shi, Sewon Min, Michihiro Yasunaga, Minjoon Seo, Rich James, Mike Lewis,
  Luke Zettlemoyer, and Wen tau Yih. 2023.
\newblock \href {https://arxiv.org/abs/2301.12652} {Replug: Retrieval-augmented
  black-box language models}.
\newblock \emph{Preprint}, arXiv:2301.12652.

\bibitem[{Song et~al.(2024)Song, Oh, Mo, Kim, Yun, Ha, and
  Shin}]{song2024hierarchicalcontextmergingbetter}
Woomin Song, Seunghyuk Oh, Sangwoo Mo, Jaehyung Kim, Sukmin Yun, Jung-Woo Ha,
  and Jinwoo Shin. 2024.
\newblock \href {https://arxiv.org/abs/2404.10308} {Hierarchical context
  merging: Better long context understanding for pre-trained llms}.
\newblock \emph{Preprint}, arXiv:2404.10308.

\bibitem[{Team(2024{\natexlab{a}})}]{Reid2024Gemini1U}
Google~Gemini Team. 2024{\natexlab{a}}.
\newblock \href {https://api.semanticscholar.org/CorpusID:268297180} {Gemini
  1.5: Unlocking multimodal understanding across millions of tokens of
  context}.
\newblock \emph{ArXiv}, abs/2403.05530.

\bibitem[{Team(2024{\natexlab{b}})}]{dubey2024llama3herdmodels}
Meta Llama~3 Team. 2024{\natexlab{b}}.
\newblock \href {https://arxiv.org/abs/2407.21783} {The llama 3 herd of
  models}.
\newblock \emph{Preprint}, arXiv:2407.21783.

\bibitem[{Wang et~al.(2023)Wang, Dong, Cheng, Liu, Yan, Gao, and
  Wei}]{wang2023augmentinglanguagemodelslongterm}
Weizhi Wang, Li~Dong, Hao Cheng, Xiaodong Liu, Xifeng Yan, Jianfeng Gao, and
  Furu Wei. 2023.
\newblock \href {https://arxiv.org/abs/2306.07174} {Augmenting language models
  with long-term memory}.
\newblock \emph{Preprint}, arXiv:2306.07174.

\bibitem[{Wu et~al.(2021)Wu, Ouyang, Ziegler, Stiennon, Lowe, Leike, and
  Christiano}]{Wu2021RecursivelySB}
Jeff Wu, Long Ouyang, Daniel~M. Ziegler, Nissan Stiennon, Ryan Lowe, Jan Leike,
  and Paul~Francis Christiano. 2021.
\newblock \href {https://api.semanticscholar.org/CorpusID:237593001}
  {Recursively summarizing books with human feedback}.
\newblock \emph{ArXiv}, abs/2109.10862.

\bibitem[{Xu et~al.(2023)Xu, Song, Iyyer, and Choi}]{xu-etal-2023-critical}
Fangyuan Xu, Yixiao Song, Mohit Iyyer, and Eunsol Choi. 2023.
\newblock \href {https://doi.org/10.18653/v1/2023.acl-long.181} {A critical
  evaluation of evaluations for long-form question answering}.
\newblock In \emph{Proceedings of the 61st Annual Meeting of the Association
  for Computational Linguistics (Volume 1: Long Papers)}, pages 3225--3245,
  Toronto, Canada. Association for Computational Linguistics.

\bibitem[{Xu et~al.(2024)Xu, Ping, Wu, McAfee, Zhu, Liu, Subramanian,
  Bakhturina, Shoeybi, and Catanzaro}]{xu2024retrievalmeetslongcontext}
Peng Xu, Wei Ping, Xianchao Wu, Lawrence McAfee, Chen Zhu, Zihan Liu, Sandeep
  Subramanian, Evelina Bakhturina, Mohammad Shoeybi, and Bryan Catanzaro. 2024.
\newblock \href {https://arxiv.org/abs/2310.03025} {Retrieval meets long
  context large language models}.
\newblock \emph{Preprint}, arXiv:2310.03025.

\bibitem[{Xu and Lapata(2022)}]{xu2022textsummarizationoracleexpectation}
Yumo Xu and Mirella Lapata. 2022.
\newblock \href {https://arxiv.org/abs/2209.12714} {Text summarization with
  oracle expectation}.
\newblock \emph{Preprint}, arXiv:2209.12714.

\bibitem[{Zha et~al.(2023)Zha, Yang, Li, and Hu}]{zha-etal-2023-alignscore}
Yuheng Zha, Yichi Yang, Ruichen Li, and Zhiting Hu. 2023.
\newblock \href {https://doi.org/10.18653/v1/2023.acl-long.634}
  {{A}lign{S}core: Evaluating factual consistency with a unified alignment
  function}.
\newblock In \emph{Proceedings of the 61st Annual Meeting of the Association
  for Computational Linguistics (Volume 1: Long Papers)}, pages 11328--11348,
  Toronto, Canada. Association for Computational Linguistics.

\bibitem[{Zhang et~al.(2023)Zhang, Wan, and
  Bansal}]{zhang-etal-2023-extractive}
Shiyue Zhang, David Wan, and Mohit Bansal. 2023.
\newblock \href {https://doi.org/10.18653/v1/2023.acl-long.120} {Extractive is
  not faithful: An investigation of broad unfaithfulness problems in extractive
  summarization}.
\newblock In \emph{Proceedings of the 61st Annual Meeting of the Association
  for Computational Linguistics (Volume 1: Long Papers)}, pages 2153--2174,
  Toronto, Canada. Association for Computational Linguistics.

\end{thebibliography}

\appendix

\section{Prompts}
\label{sec:prompts}

Table~\ref{tab:prompts_first_lvl} contains  first-level prompt templates,  for abstractive summaries and summaries with citations. Table~\ref{tab:prompt_sub_level_hm} shows subsequent-level templates for the original hierarchical merging (HMerge) and our adaptation with citations.  Table~\ref{tab:prompt_sub_level_support} presents   subsequent-level templates for Extract-Support, Retrieve-Support, and Cite-Support.

\begin{table}[h]
\begin{tcolorbox}[colback=white, colframe=blue1, left=2pt,  coltitle=white, halign title=flush center, title=\textbf{First-level Chunk Summary}, halign=justify]

    \texttt{Below is a document:} \\
    {-}{-}{-} \\
    \texttt{\{\}} \\
    {-}{-}{-} \\
    \texttt{Write a summary containing all key information. There should be no explicit mention of words like "document" or "summary" in the summary.} \\
    \end{tcolorbox}
    
    \begin{tcolorbox}[colback=white, colframe=blue1, left=2pt,  halign title=flush center, coltitle=white, title=\textbf{First-level Chunk Summary w/Citations}, halign=justify]
    
    \texttt{Below is a document with each paragraph assigned to a label at the end ([n]) and separated by line breaks:} \\
    {-}{-}{-} \\
    \texttt{\{\}} \\
    {-}{-}{-} \\
    \texttt{Write a summary containing all key information. There should be no explicit mention of words like "document" or "summary" in the summary. After each summary sentence, you should assign a label to that sentence showing which paragraph in the document it corresponds to. Specifically, follow the format below:} \\

    \texttt{<sentence 1>. [n] <sentence 2>. [m] ...} \\
  \end{tcolorbox}
 \caption{First-level prompt templates for generating intermediate summaries without and with citations.}
 \label{tab:prompts_first_lvl}
\end{table}

\begin{table}[h!]
\begin{tcolorbox}[colback=white, colframe=blue2, left=2pt,  halign title=flush center, coltitle=white, title=\textbf{Subsequent-level Merging:  HMerge}, halign=justify]
  
  \texttt{Below are several summaries of different parts of a document:} \\
    {-}{-}{-} \\
    \texttt{\{\}} \\
    {-}{-}{-} \\
    \texttt{Merge the given summaries into one single summary containing all key information. There should be no explicit mention of words like "document" and "summary" in the summary.} \\
\end{tcolorbox}

\begin{tcolorbox}[colback=white, colframe=blue2, left=2pt,  halign title=flush center, coltitle=white, title=\textbf{Subsequent-level Merging: CiteHMerge}, halign=justify]
   
 \texttt{Below are several summaries of different parts of a document with each summary having sentences with labels at the end ([1], [2], ...) and separated by line breaks:} \\
    {-}{-}{-} \\
    \texttt{\{\}} \\
    {-}{-}{-} \\
    \texttt{Merge the given summaries into one single summary containing all key information. There should be no explicit mention of words like "document" and "summary" in the summary. After each summary sentence, you should assign a label to that sentence showing which paragraph in the document it corresponds to. Specifically, follow the format below:} \\ \\ 
    \texttt{<sentence 1>. [n] <sentence 2>. [m] ...} 
  \end{tcolorbox}
 \caption{Subsequent-level prompt templates for Original-HM and Cite-HM.}
  \label{tab:prompt_sub_level_hm}
\end{table}

\begin{table*}
\begin{tcolorbox}[colback=white, colframe=blue3, left=2pt,  coltitle=white, halign title=flush center, title=\textbf{Subsequent-level Merging: Extract/Retrieve Support}, halign=justify]
    \texttt{Below are several summaries of different parts of a document:} \\
    {-}{-}{-} \\
    \texttt{\{\}} \\
    {-}{-}{-} \\
    \texttt{Below are the supporting contexts of the previously shown summaries:} \\
    {-}{-}{-} \\
    \texttt{\{\}} \\
    {-}{-}{-} \\
    \texttt{Merge the given summaries into one single summary containing all key information and use the supporting contexts to make sure the merged summary contains no factual errors. The gist of the summary should be based solely on the given summaries, while the supporting contexts should be used for proofreading only. There should be no explicit mention of words like "document", "context" or "summary" in the summary.} \\
\end{tcolorbox}
\begin{tcolorbox}[colback=white, colframe=blue3, left=2pt,  coltitle=white, halign title=flush center, title=\textbf{Subsequent-level Merging: Cite-Support}, halign=justify]
    \texttt{Below are several summaries of different parts of a document:}\\
    {-}{-}{-} \\
    \texttt{\{\}} \\
    {-}{-}{-} \\
    \texttt{Below are the supporting contexts of the previously shown summaries, with each context assigned to a label at the end ([n]) and separated by line breaks:} \\
    {-}{-}{-} \\
    \texttt{\{\}} \\
    {-}{-}{-} \\
    \texttt{Merge the given summaries into one single summary containing all key information and use the supporting contexts to make sure the merged summary contains no factual errors. The gist of the summary should be based solely on the given summaries, while the supporting contexts should be used for proofreading only. There should be no explicit mention of words like "document", "context" or "summary" in the summary. After each summary sentence, you should assign a label to that sentence showing which supporting context it corresponds to. Specifically, follow the format below:} \\ \\ 
    \texttt{<sentence 1>. [n] <sentence 2>. [m] ...} \\
  \end{tcolorbox}
  \caption{Subsequent-level prompt templates for Support methods.}
  \label{tab:prompt_sub_level_support}
\end{table*}

\section{Algorithm for Ranking Citations}
\label{sec:alg_cite}

\begin{algorithm*}
\caption{Passage Selection with Coverage Constraints}
\label{alg:attr_select}
\begin{algorithmic}
    \State attrTexts $\leftarrow$ List of input passages
    \State response $\leftarrow$ The summary w/ citations
    \State k $\leftarrow$ No. of passages to select \\

    \State allLabels $\leftarrow$ \textsc{ExtractLabels}(response)
    \State labelCounts $\leftarrow$ \textsc{CountLabelFrequencies}(allLabels) \Comment{dict in format \{label: count\}}
    \State \textsc{Sort}(labelCounts, key=count, order=descending) \\

    \State selectedPassages $\leftarrow$ []
    \State labelToPick $\leftarrow$ None
    \ForAll{(label, count) $\in$ labelCounts}
      \If{count + \textsc{Len}(selectedPassages) $>$ k}
        \State labelToSelect $\leftarrow$ label
        \State break
      \Else
        \State selectedPassages += \textsc{GetPassagesByLabel}(attrTexts, label)
      \EndIf
    \EndFor \\

    \If{labelToSelect == None}
      \State return selectedPassages
    \Else
      \State remainingPassages $\leftarrow$ \textsc{GetPassagesByLabel}(attrTexts, labelToPick)
      \State remainingPassageIndices $\leftarrow$ \textsc{IndexOf}(remainingPassages, attrTexts)
    \EndIf \\

    \State converageSections $\leftarrow$ $[\frac{2i + 1}{2k}\,\text{for i} \in 0 \cdots k - 1]$
  
    \ForAll{passage $\in$ selectedPassages}
      \State pos $\leftarrow$ \textsc{IndexOf}(passage, attrTexts)
      \State closestSection $\leftarrow$ \textsc{FindClosestSection}(coverageSections, pos)
      \State coverageSections $\leftarrow$ coverageSections $\setminus$ \{closestSection\}
    \EndFor \\

    \ForAll{passage, passageIndex $\in$ remainingPassage, remainingPassageIndices}
      \State closestSection $\leftarrow$ \textsc{FindClosestSection}(coverageSections, passageIndex)
      \State distToClosestSection $\leftarrow$ \textsc{DistToClosestSection}(coverageSections, passageIndex)
      \If{distToClosestSection is minimum}
        \State selectedPassages $\leftarrow$ selectedPassages $\cup$ \{passage\}
        \State coverageSections $\leftarrow$ coverageSections $\setminus$ \{closestSection\}
      \EndIf
      \If{\textsc{Len}(selectedPassages) == k}
        \State return selectedPassages
      \EndIf
    \EndFor \\

\end{algorithmic}
\end{algorithm*}

Algorithm~\ref{alg:attr_select} describes how we select passages based on citation labels extracted from the summary. First, we build a dictionary with each  label being the key, and the number of times it appears in the summary its value. We then append all passages that belong to the most frequent label, remove this label from the dictionary, and repeat this process until  all passages are appended or number of passages to be selected (denoted as $k$) is exceeded. If no passages are left or we have appended exactly $k$ passages, we return the selected passages. Otherwise, we select passages for the next label to fill the remaining vacancies.

When filling the remaining vacancies, we aim to maximize the coverage of the input. We first divide the input text into $k-1$ sections of equal length, so that ideally the selected passages should cover all sections. For each passage that has been  already selected, we find its closest section based on its position in the input text and remove that section from consideration. Then, for each remaining candidate passage (those belonging to the next most frequent label), we calculate its distance to each remaining section. We iteratively select passages that have the minimum distance to any remaining uncovered section, removing that section from consideration after each selection. This process continues until we have selected exactly~$k$ passages in total. Overall, This approach ensures that the selected passages not only represent the most frequently cited content (through label frequency) but also provide broad coverage across different portions of the source text, avoiding redundant selection from the same regions.

\section{Additional Results on Fact-Precision and Fact-Recall for \textsc{PRisma}}
\label{sec:fact_prec_rec}

Tables~\ref{tab:8b_extra} and \ref{tab:70b_extra} present  results for Fact-Precision and Fact-Recall for the  8B and 70B Llama models, respectively. In general, we see that Extract-Support has the best overall performance, achieving the highest scores most frequently. Retrieve-Support trails closely behind with multiple highest and second highest appearances. We also observe that the original HMerge method shows strong performance at \textsc{Fact-Prec}, but falls behind on \textsc{Fact-Rec} which leads to worse \textsc{PRisma} scores.

\begin{table}[t]
  \centering
  \renewcommand{\arraystretch}{1.5}
  \fontsize{9}{10}\selectfont
  \begin{tabular}{lcc}
    \hline
    \large{\textbf{Multi-LexSum}} & \textsc{Fact-Prec} & \textsc{Fact-Rec} \\
    \hline
    Zero-shot & 50.3 & 37.1 \\
    HMerge & 55.1 & 38.8 \\
    Cite-HMerge & 51.7 & 36.8 \\
    \hline
    Retrieve-Replace & \uline{56.1} & 33.3 \\
    Retrieve-Support & 55.2 & \uline{39.6} \\
    \hline
    Extract-Replace & 48.2 & 37.1 \\
    Extract-Support & \textbf{56.3} & \textbf{41.1} \\
    \hline
    Cite-Replace & 40.8 & 32.7 \\
    Cite-Support & 50.4 & 37.5 \\
    \hline
    \multicolumn{3}{c}{} \\ \hline
    \large{\textbf{SuperSummary}} & \textsc{Fact-Prec} & \textsc{Fact-Rec} \\
    \hline
    Zero-shot & \uline{60.1} & 22.9 \\
    HMerge & \textbf{61.5} & 27.3 \\
    Cite-HMerge & 45.6 & 28.9 \\
    \hline
    Retrieve-Replace & 29.0 & 19.5 \\
    Retrieve-Support & 54.5 & \uline{29.6} \\
    \hline
    Extract-Replace & 33.8 & 16.7 \\
    Extract-Support & 56.0 & \textbf{30.1} \\
    \hline
    Cite-Replace & 29.2 & 15.6 \\
    Cite-Support & 46.4 & 27.7 \\
    \hline
  \end{tabular}
  \caption{\textsc{Fact-Prec} and \textsc{Fact-Rec} Results for the Llama-3.1-8B-Instruct model. \textsc{Fact-Prec} and \textsc{Fact-Rec} refer to fact precision and recall obtained while computing PRISMA. Best results for each metric are highlighted in \textbf{bold}, while the second best results for each metric are \uline{underlined}.}
  \label{tab:8b_extra}
\end{table}

\begin{table}[!t]
  \centering
  \renewcommand{\arraystretch}{1.5}
  \fontsize{9}{10}\selectfont
  \begin{tabular}{lcc}
    \hline
    \large{\textbf{Multi-LexSum}} & \textsc{Fact-Prec} & \textsc{Fact-Rec} \\
    \hline
    Zero-shot & 46.7 & 37.3 \\
    HMerge & \uline{57.5} & 41.5 \\
    Cite-HMerge & 54.9 & 41.6 \\
    \hline
    Retrieve-Replace & 57.1 & 35.6 \\
    Retrieve-Support & 57.1 & \textbf{44.1} \\
    \hline
    Extract-Replace & 50.6 & 36.9 \\
    Extract-Support & \textbf{57.8} & \uline{43.7} \\
    \hline
    Cite-Replace & 46.7 & 35.9 \\
    Cite-Support & 54.6 & 41.0 \\
    \hline
    \multicolumn{3}{c}{} \\ \hline 
    \large{\textbf{SuperSummary}} & \textsc{Fact-Prec} & \textsc{Fact-Rec} \\
    \hline
    Zero-shot & 55.8 & 25.8 \\
    HMerge & \uline{59.4} & 32.7 \\
    Cite-HMerge & 49.9 & 31.7 \\
    \hline
    Retrieve-Replace & 40.7 & 18.5 \\
    Retrieve-Support & \textbf{60.6} & \uline{34.9} \\
    \hline
    Extract-Replace & 45.1 & 15.9 \\
    Extract-Support & 57.2 & \textbf{37.9} \\
    \hline
    Cite-Replace & 21.5 & 22.1 \\
    Cite-Support & 52.7 & 33.3 \\
    \hline
  \end{tabular}
  \caption{\textsc{Fact-Prec} and \textsc{Fact-Rec} Results for the Llama-3.1-70B-Instruct (GPTQ-INT4) model. \textsc{Fact-Prec} and \textsc{Fact-Rec} refer to fact precision and recall obtained while computing PRISMA. Best results for each metric are highlighted in \textbf{bold}, while the second best results for each metric are \uline{underlined}.}
  \label{tab:70b_extra}
\end{table}

\section{Examples of Model Output}
\label{sec:example_summaries}

This section contains output summaries for different models for the book \textit{Things We Hide from the Light} (by Lucy Score). Table~\ref{tab:empsumm_reference} contains the Reference summary, whereas Tables~\ref{tab:empsumm_zs}, \ref{tab:empsumm_original_hm}, \ref{tab:empsumm_cite_hm}, \ref{tab:empsumm_ext_sup}, and \ref{tab:empsumm_ext_ret} respectively present summaries corresponding to the  Zero-shot model, the original HMerge method, Cite-HMerge our adaptation of HMerge with citations,  and the proposed Extract-Support summary and Extract-Retrieve methods.

\begin{table*}
\begin{tcolorbox}[colback=white, colframe=violet, left=2pt,  coltitle=white, halign title=flush center, title=\textbf{Reference Summary}, halign=justify]
\begin{small}
It has been months since Nash Morgan, Police Chief of Knockemout, Virginia, was shot attempting to break up a stolen car ring run by Duncan Hugo, the son of an underworld kingpin in nearby Washington. Even now, FBI special agents shadow Nash to protect him against a possible reprisal by the crime family. Nash has post-traumatic stress disorder (PTSD) and cannot recall the shooting. Despite counseling and anxiety medication, Nash feels isolated and paranoid, and is prone to anxiety attacks. 
\\

Angelina Solavita, an insurance investigator known for tracking down stolen property, moves into the apartment next door to Nash. She is the ex-lover of Nashs older brother, Knox. Wrestling with her own kind of PTSD (she nearly died from a heart defect when she was 15), Lina hesitates to share with Nash that she is in town to recover a vintage Porsche stolen by the same crime ring responsible for his shooting. Although they are immediately attracted to each other, Lina sees the hunky Nash more as a fling given her nomadic lifestyle. Meanwhile, Nash, who has trauma because of the shooting, yearns for the stability of a relationship. \\

At work, Nash deals with Officer Tate Dilton, a rookie cop who has earned a string of harassment complaints. The department is in legal trouble after Dilton pulled over a Black couple driving a luxury car on the false suspicion (prompted by racial profiling) that the car must be stolen. The couple is now contemplating suing the department. Nash puts Dilton on paid leave. \\

Lina gathers evidence about the missing Porsche. As she digs, she begins to suspect the thug who shot Nash may be part of the same crime organization that stole the sports car. As Nashs and Linas investigations progress, the two act on their attraction. They share a passionate kiss, flirt, and, when Nash experiences a panic attack, they spend a chaste night together. Gradually, they open up to each other about their near-death experiences and their feelings of vulnerability. Lina tells Nash why she is in Knockemout. When the two finally have sex, it is life-altering in its intimacy and its intensity. Lina takes a reluctant Nash skydiving, and as the two fall to earth harnessed to each other, they both feel the profound pull of love.\\

When a drunken Tate Dilton and his beer buddies harass Lina and her friends during a town Halloween party, Nash intervenes alongside his brother and several friends. A fight breaks out. The following morning, Nash relieves Tate of his duties.\\

Using a network of friends, Lina and Nash coordinate their efforts to identify who shot Nash. Solving that mystery will lead them to Duncan Hugo and the stolen Porsche. On a hunch from Waylay, Nashs 12-year-old niece who was abducted by the car thieves the night Nash was shot, Lina turns her attention to identifying a shadowy man who has been watching her. Waylay recalls that on the night of Nashs shooting, one of the thugs ate wrapped candy; Lina notices that the man she keeps seeing purchases bags of the same candy at a convenience store.\\

Nashs network of friends digging into the Hugo criminal organization learns that Duncan wants to cooperate with federal agents in return for eliminating his own father so that he can take over the family operation. Duncan has never left Knockemout, which means that Nashs shooting was a botched effort by an unnamed underling, but the question of who and where Duncan is hiding furthers the mystery. \\

One day, after rushing to the scene of nasty car accident, Nash encounters his estranged father, who is recovering from misusing alcohol after Nashs mother died. Nash feels overwhelmed, helpless, and unable to protect anybody. Without telling her the reason, Nash breaks it off with Lina. \\

One day, two thugsone of them the candy mansnatch Lina and drive her to a barn at an abandoned horse farm outside of Knockemout. Lina is stunned to be met by Tate Dilton, who turns out to be one of Duncan Hugos henchmen.\\

Duncan himself appears. Ever resourceful, Lina slips her restraints and surprises Duncan and Tate as they play a video game. She snatches the console, eludes them both, and hides in one of the stalls in the barn. On a hunch, she uses the consoles IP signal to contact Waylay, who in turn contacts Nash. Within minutes, Nash and his friends swarm the farm. Nash suddenly remembers that it was Tate Dilton who shot him that night. In a tense showdown, Tate pulls his gun, and then Nash shoots him dead. Duncan Hugo is apprehended. Lina and Nash fall into each others arms. \\

Days later, at his brothers wedding, Nash and Lina commit to each other. Nash gives Lina an engagement ring, and the two talk happily about the future they will have together. Five years later, Lina and Nash, now married, await the birth of twins. \\ 
\end{small}
\end{tcolorbox}
\caption{Reference summary.}
\label{tab:empsumm_reference}
\end{table*}

\begin{table*}
\begin{tcolorbox}[colback=white, colframe=violet, left=2pt,  coltitle=white, halign title=flush center, title=\textbf{Zero-shot Summary}, halign=justify]
\begin{small}
Lina Solavita, an insurance investigator, moves to Knockemout, a small town in northern Virginia, to find a 1948 Porsche 356 convertible that was stolen from a client. She is staying in an apartment next to Nash Morgan, the chief of police, who is still recovering from a shooting that left him with bullet wounds and memory loss. \\

Nash is struggling to cope with his new reality and is haunted by the sound of crunching and a sizzle or crunch that echoes the moment he was shot. Lina is initially attracted to Nash, but their relationship is complicated by his vulnerability and her job, which requires her to keep secrets. She is also dealing with her own emotional baggage, including a heart condition that almost killed her at 15. \\

Meanwhile, a U.S. marshal, Nolan Graham, is watching Nash due to a threat, and Lina is caught in the middle. As Lina and Nash spend more time together, they develop a strong connection, but their differences and secrets threaten to tear them apart. Lina's past is slowly revealed, including her history with Nolan and her job as an insurance investigator. Nash is torn between his growing feelings for Lina and his need for honesty and trust. \\

The story takes a dark turn when Lina's investigation puts her in danger, and Nash is forced to confront his own vulnerabilities. The two are eventually kidnapped by a group of bikers, who are revealed to be working for Grim, a motorcycle club leader.\ Grim reveals that Duncan Hugo, the man who shot Nash, is still in town and planning to take over his father's criminal empire. \\

As the story unfolds, Lina and Nash's relationship deepens, and they work together to uncover the truth about Duncan Hugo's plans. They also face challenges from Tate Dilton, a disgraced police officer, and Lucian Rollins, Nash's friend, who is hiding secrets of his own. 

The story is full of twists and turns, including a dramatic confrontation at the library's Book or Treat event, where Lina and her friends are attacked by a group of bikers. \\

Throughout the story, Lina and Nash's relationship is a central theme, and their chemistry is undeniable. However, their secrets and lies threaten to destroy their trust and ultimately their relationship. The story is a complex exploration of love, trust, and vulnerability, set against the backdrop of a small town with its own secrets and mysteries. \\

The story is also filled with a cast of colorful characters, including Mrs. Tweedy, a feisty elderly woman who is always up for a fight, and Naomi, a kind-hearted community outreach coordinator who becomes Lina's friend. \\

The story is a gripping and emotional ride, with a complex plot that keeps readers guessing until the end. \\ 
\end{small}
\end{tcolorbox}
\caption{Zero-shot summary.}
\label{tab:empsumm_zs}
\end{table*}

\begin{table*}
\begin{tcolorbox}[colback=white, colframe=violet, left=2pt,  coltitle=white, halign title=flush center, title=\textbf{Original HMerge Summary}, halign=justify]
\begin{scriptsize}
Lina Solavita, an investigator for Pritzger Insurance, moved to Knockemout, Virginia, where she became entangled in a complicated situation involving Nash Morgan, the police chief. Nash was struggling to come to terms with his new reality after being shot in the line of duty and was feeling disconnected from his life. Lina's presence sparked a series of events, including a confrontation with US Marshal Nolan Graham, who was shadowing Nash. Lina's past was marked by a medical condition and a complicated relationship with physical touch, which made her interactions with Nash tense and awkward. \\

As Lina settled into her new apartment, she received a call from her mother, Bonnie, who was checking in on her. Lina reassured her that she was fine and that she was enjoying her time in Knockemout. However, Lina's mother was not convinced, and she was already making plans to visit her daughter soon. Lina's conversation with her mother was interrupted by a knock at the door. It was Knox Morgan, Nash's brother, who was there to talk to Nash about something. Nash was not interested in talking, but Knox was insistent. As Knox headed to Nash's apartment, Lina was left to wonder what was going on between the two brothers. \\

Despite the challenges she was facing, Lina was determined to make the most of her time in Knockemout. She started to feel a connection with Nash, and she was hoping to help him heal and find his way again. As she looked around her new apartment, Lina felt a sense of possibility and promise. She was not sure what the future held, but she was ready to face whatever came her way. \\

Nash was struggling to come to terms with his past and his feelings for Lina. He was haunted by nightmares and was trying to find a way to heal and move forward. Lina was a strong and independent character, but she was also vulnerable and guarded. Nash was a complex character with a troubled past, but he was also kind and heroic. The chemistry between them was palpable, and their interactions were filled with tension and humor. As Lina navigated her feelings for Nash, she must confront her own vulnerabilities and learn to trust him. Nash's brother, Knox, was a constant presence in the background, offering advice and guidance. However, Nash was determined to forge his own path and make his own decisions. The investigation into the police officer's misconduct and the mysterious henchmen continued to unfold, and it was unclear what the outcome would be. Lina's friends, Stef, Naomi, and Sloane, were trying to support her through her struggles. They were also dealing with their own personal issues, including a recent abduction and a complicated relationship with Nash's brother, Knox. 
In a separate storyline, Nolan, a marshal, was trying to get Lina to open up about her past. He was a supportive friend and was trying to help her heal and move forward. The two had a non-date drink together, and Lina began to open up about her feelings and experiences.\\

As the story unfolded, it became clear that Nash and Lina's relationship was complex and multifaceted. They were drawn to each other, but their past experiences and personal issues made it difficult for them to be together. The investigation into the police officer's misconduct and the mysterious henchmen continued to unfold, and it was unclear what the outcome would be. 
Lina was kidnapped by Cereal Aisle Guy, who was revealed to be Nikos, and taken to a barn-like building where she met Tate Dilton, a dirty cop, and Duncan Hugo, the mastermind. Hugo revealed that Dilton was hired to shoot Nash, but instead shot him in cold blood on the highway. Hugo planned to use Lina to lure Nash to the barn, where he would be killed. Lina tried to stall and gather information, learning that Hugo's goal was to take over his father's business and that Dilton had a personal vendetta against Nash.\\

Nash and his friends were searching for Lina, following a lead on a partial plate number that Waylay memorized. They were coordinating with the state police and setting up traffic stops to find Lina and the unsub. In a desperate attempt to escape, Lina used a microphone to call for help, reaching out to a group of children playing an online video game. The kids, including a girl named Brecklin, were initially skeptical but eventually believed Lina's plea for help. One of the players, Waylay, recognized Lina's name and realized that her uncle, Nash, was the chief of police. Waylay joined the game and helped Lina by providing crucial information to Nash. Meanwhile, Nash was searching for Lina and had tracked down a lead to a Ford Fusion belonging to Mark Nikos. \\

Nash, along with his brother Knox and friend Nolan, approached the farm, where they found Lina being held by Dilton and Hugo. A shootout ensued, and Lina managed to escape by using a pressure washer to defend herself. She then stabbed Hugo with a pitchfork and made her way to a Porsche, where she met up with Knox and Nolan. As they tried to escape, Dilton shot Nolan, and Lina and Knox took cover behind a tractor. Nash arrived and proposed to Lina, who accepted. Knox then took Lina and Nolan to safety, while Nash stayed behind to confront Dilton. Nash was able to outmaneuver Dilton and shot him three times, killing him. \\

The story then cut to a wedding ceremony, where Knox was marrying Naomi. Lina was in attendance, and Nash was serving as a groomsman. The ceremony was emotional, with many guests in tears. After the ceremony, Nash and Lina shared a dance, and he told her that he'd been thinking about their future together. He revealed that he got a tattoo of angel wings on his butt, which Lina found romantic.\\ 
\vspace{-.1cm}

As they danced, Nash's father, Duke, approached them and congratulated them on their engagement. Duke was sober and seemed to be making amends for his past mistakes. Nash and Lina then sneaked away to Nash's old bedroom, where they shared a romantic moment. However, their intimacy was interrupted by Lina's parents, who had arrived unexpectedly. Nash was caught off guard, but Lina was relieved that her parents had finally met Nash. \\ \vspace{-.4cm}
\end{scriptsize}
\end{tcolorbox}
\vspace{-.4cm}
\caption{Original HMerge Summary.}
\label{tab:empsumm_original_hm}
\end{table*}

\begin{table*}
\begin{tcolorbox}[colback=white, colframe=violet, left=2pt,  coltitle=white, halign title=flush center, title=\textbf{Cite-HMerge Summary}, halign=justify]
\begin{small}
Nash Morgan, a police chief, is struggling to cope with the emotional aftermath of being shot in the line of duty, and his emotional numbness is affecting his relationships with those around him. He meets his new neighbor, Angelina Solavita, also known as Lina, who is an insurance agent and his brother's ex-girlfriend from college, and despite initial reservations, they are drawn to each other. \\

Lina is investigating a case involving a stolen car, and her search leads her to cross paths with Nash, who is still recovering from his injuries and struggling to come to terms with his feelings about her. As they spend more time together, they grow closer, but their relationship is complicated by Nash's past trauma and Lina's secrets. Lina is searching for a car that Duncan Hugo stole and is looking for any leads in the crime scene files. She is also being pursued by Nash, who is concerned about her safety and thinks she should not be involved in the investigation. \\

Lina observes a fight between two brothers, Wendell Baker and his brother, and calls 911, but before the police arrive, she is abducted by two people in a van. They take her to meet Grim, the leader of a motorcycle club, who has information about Hugo. Grim reveals that Hugo is still in town and is planning to take over his father's crime business. Lina is frustrated that she is not being included in the plans to deal with Hugo and decides to take matters into her own hands. \\ 

She attends a Halloween party at the library, where she runs into an old friend from high school, Angie, and her family. Angie's son, Austin, is recovering from leukemia, and Angie has been thinking about how she can be a better parent to him. Lina and Angie reconnect, and Angie apologizes again for not being a better friend. Lina reflects on how she pushed her friends away after her own health issues and how she is grateful for Angie's apology. \\

Nash and his friends are preparing for a confrontation with Hugo and his men. Lina is trying to stay one step ahead of Hugo and his plans. She is also dealing with her feelings for Nash and trying to navigate their complicated relationship. \\

Angie's husband is supportive of her reconnecting with Lina and is grateful for her strength and resilience. Lina and Angie are rebuilding their friendship and supporting each other. \\

The story is ongoing, with Lina and Nash working together to take down Hugo and his men. Lina is a strong and determined protagonist who is not afraid to take risks. She is also vulnerable and relatable, with a deep emotional life. The story is a thrilling and romantic ride, with many surprises along the way. Lina and Nash's relationship is complicated and passionate, with a deep emotional connection. The story is full of suspense and action, with many twists and turns.\\

Angelina, dressed as Nancy Drew, is at a Halloween party with her friends Naomi and Sloane, when they are approached by Nash Morgan and his brothers, Knox, Nolan, and Lucian. The group is interrupted by a drunk man who grabs Sloane, leading to a confrontation that ends with Nash arresting the man and his friend, who had also gotten into a fight with Nolan. After the altercation, Nash asks Angelina to dance, and despite initial hesitation, she agrees. As they dance, the tension between them becomes palpable, and Nash eventually pulls her away to a quiet room, where they share a passionate kiss. \\

Angelina had a life-changing experience with Nash, which left her questioning her feelings and the connection they shared. The night before, Nash had been on his knees, pleasuring her, and she had experienced multiple orgasms, which was a new and overwhelming experience for her. After their intense encounter, Nash was still confident and self-assured, which only added to Angelina's confusion and frustration. Nash asked her what she needed to feel safe with him, and Angelina struggled to articulate her feelings, unsure of how to process the intensity of their connection. Nash and Angelina made a deal to work together and be open with each other. Nash later reflected on their encounter and realized that he had been given a second chance with Angelina. He sent her a playful text message, teasing her about her exhaustion and inviting her to dinner. Angelina responded, and they exchanged banter, with Nash inviting her to dinner and Angelina agreeing. Nash also had a conversation with Sloane, who had discovered that he and Angelina had slept together. \\

Nash and Angelina's dinner date ended on a romantic note, with Nash and Angelina feeling closer and more connected. Angelina revealed that she wanted to be part of the team working to catch Duncan Hugo and help keep Nash safe. Nash was touched by her concern and willingness to help. He realized that Angelina cared about him deeply and was willing to take risks to protect him. \\ 
\end{small}
\end{tcolorbox}
\caption{Cite-HMerge Summary.}
\label{tab:empsumm_cite_hm}
\end{table*}

\begin{table*}
\begin{tcolorbox}[colback=white, colframe=violet, left=2pt,  coltitle=white, halign title=flush center, title=\textbf{Extract-Support Summary}, halign=justify]
\begin{small}
In the small town of Knockemout, Police Chief Nash Morgan is struggling to cope with the aftermath of being shot in the line of duty. He's emotionally numb and feeling useless, going through the motions of his job without any real passion or interest. His life is further complicated by the arrival of his brother's ex-girlfriend, Angelina "Lina" Solavita, who moves into the apartment next door. \\

Lina is a beautiful and charming woman who works as an insurance investigator, and Nash finds himself drawn to her despite his emotional numbness. Lina is in town investigating a case involving a list of law enforcement officers and informants, including Nash, who was targeted by Duncan Hugo, the son of a wealthy and influential man. She's also on the hunt for a stolen 1948 Porsche 356 convertible, which is priceless to her client. \\

As Lina navigates her investigation, she befriends Naomi, Knox's fiancée, and Sloane, a librarian, who support her and help her through her tough time. Nash, still recovering from his injuries and struggling with memory loss and panic attacks, is haunted by the memory of the shooting and is experiencing flashbacks. Lina is a calm and soothing presence in his life, and he finds himself feeling more at ease around her. \\

As Nash and Lina navigate their growing attraction, they must also confront the secrets and mysteries that surround them. Lina's investigation leads her to a row home, where she observes a confrontation between two brothers, Wendell Baker and his brother. The situation escalates, and Lina is abducted by a group of people in ski masks, who turn out to be associates of a man named Grim. Grim is a leader of a motorcycle club and has information about Duncan Hugo's whereabouts. \\

As Nash and Lina's relationship deepens, they grow closer, and Lina begins to confront her fears and consider the possibility of opening up to Nash and her parents. The investigation into Duncan Hugo's whereabouts continues, with the team working together to gather information and stay safe. The team comes up with a plan to use Nash as bait to lure Hugo out of hiding. \\

Lucian Rollins, a wealthy and influential man, is concerned about Lina's intentions and warns her to stay away from Nash. However, Lina is not intimidated and fires back at Lucian, accusing him of trying to manipulate her and warning him to respect Nash's boundaries. Despite the tension between them, Lucian eventually becomes an ally to Nash and Lina, and he helps them in their quest to bring Hugo to justice. \\

As the team prepares for the showdown with Hugo, tensions are running high. Knox is worried about the safety of his fiancée, Naomi, and her daughter, Waylay, while Lucian is concerned about the potential risks of the plan. Despite these concerns, the team is determined to bring Hugo to justice. \\

In the end, Nash and Lina's relationship is put to the test as they work together to take down Hugo and bring him to justice. They face challenges and obstacles, but their connection remains strong. As they navigate the ups and downs of life, they realize that their love is the key to overcoming any obstacle. Nash proposes to Lina in the midst of the chaos, and she accepts. \\

The story concludes with Nash and Lina emerging from a chaotic and intense situation, surrounded by police and other responders, as a new day begins. Throughout the story, Nash struggles with his emotions after a traumatic event where a woman was severely injured in a car accident. He feels guilty and helpless, remembering the death of his mother in a similar accident. He pushes away his girlfriend, Lina, saying he needs space and doesn't want her around anymore. However, Lina comforts Nash, telling him that he is a hero for facing his demons every day to protect his town and its people. \\

The story also explores Nash's complicated relationship with his father, Duke, who has been absent for a while but is making an effort to reconnect with his sons. Nash is wary at first, but he's happy to see his father making progress and is grateful for his presence at Knox's wedding. \\ 
\end{small}
\end{tcolorbox}
\caption{Extract-Support Summary.}
\label{tab:empsumm_ext_sup}
\end{table*}

\begin{table*}
\begin{tcolorbox}[colback=white, colframe=violet, left=2pt,  coltitle=white, halign title=flush center, title=\textbf{Extract-Retrieve Summary}, halign=justify]
\begin{small}
Sloane, a former sophomore, and Lucian, a former senior, were next-door neighbors. Sloane had bought her house from her parents when they moved, while Lucian kept his mom's place. On one occasion, Lou and Amanda volunteered to walk Liza J home, sparking a discussion about who would foster her. \\

Anthony Hugo, a notorious crime lord, controlled a vast territory that included Washington, DC, and Baltimore. Naomi and Sloane simultaneously expressed their interest in fostering Liza J. Knox and Lucian stood alongside Nash and Nolan, while Officer Dilton revealed that Mr. Peters was Knockemout's solicitor. Lucian proposed a theory that Anthony had helped his son leave the country. \\

The community had come together to organize a party, with library patrons buying tickets and business sponsors, persuaded by Sloane, donating food and drinks. The library retained the profits from the event. Special Agent Idler had a personal vendetta against Anthony Hugo and was determined to bring him to justice. \\

Naomi and Knox, along with their extended families, were instrumental in maintaining order in the community. However, Nash Morgan posed a significant threat to women everywhere. Naomi's soon-to-be husband was reminded to take a time-out as Naomi, Sloane, and Mrs. Tweedy approached. \\

Tallulah and Justice St. John, along with Gael and his husband, Isaac, occupied a table, enjoying their monthly double date. They were seated next to Fi and her husband. Naomi and Sloane were taking their assignments seriously, and Sloane raised her hand to contribute to the discussion. According to the Knockemout grapevine, Knox, Nolan, Lucian, and others had gathered in Knox's secret lair office. Knox's gaze met his brother's, and they shared a long, meaningful look. Jeremy Trent, the former captain of the baseball team, had beaten out Dilton for homecoming king during their senior year of high school. \\

Knox drove Waylay and Liza J home to Naomi, who was tracked as being at home. Naomi and Sloane also approached with a proposal, inquiring if they would be interested in helping with their new venture. Naomi's wedding was scheduled for the next day, and Waylay was just a child. Lina was spotted in the photo booth again with Sloane and Fi, as they celebrated the upcoming wedding. The atmosphere was filled with discussions of real estate, marriage, and babies. \\ 

Liza J's presence was a reminder of the challenges that lay ahead. Lucian and a private security team were tasked with guarding them, ensuring their safety. Lucian was seen clutching his prized Smith \& Wesson six-shooter, a symbol of his commitment to protecting those he cared about. He nodded and followed Melissa inside, as they worked together to address the looming threat. Duncan Hugo was wanted for attempted murder of a law enforcement officer, domestic violence, and assault. \\

Concerns about Naomi's safety were raised, and Lucian reassured everyone that she was being protected. Hugo's careless attitude was a stark contrast to the gravity of the situation. A warning was issued to call Nash and inform him that Duncan Hugo was sending Tate Dilton to their house. The situation was dire, and it was essential to take immediate action. The challenge of explaining the situation to a twelve-year-old was daunting. Duncan Hugo was in custody, but the threat to their safety remained. \\

A plan was devised to fly in Liza J's family to surprise her, a gesture that would bring some joy amidst the chaos. The family tradition of grand gestures was evident, as Lou recalled surprising Mandy with a three-week cruise. Knox, the groom, was the focus of attention, and it was essential to ensure his safety, as well as that of Nolan and Liza J. Jeremiah and Lou asked in unison if they should take them to a safe location. \\ 

The atmosphere was tense, with the sound of a Ford Fusion, belonging to Mark Nikos, a reminder of the outside world. As the events unfolded, it became clear that the community was coming together to support each other. Naomi and Sloane's commitment to their assignments was evident, and their dedication to the community was inspiring. The presence of Lucian and the private security team provided a sense of reassurance, as they worked together to address the threats posed by Duncan Hugo and Nash Morgan. The celebration of Naomi's wedding was a beacon of hope, a reminder that even in the face of adversity, the community could come together and find joy. \\ 
\end{small}
\end{tcolorbox}
\caption{Extract-Retrieve Summary.}
\label{tab:empsumm_ext_ret}
\end{table*}

\section{Manual Analysis of Summary Facts}
\label{sec:facts_verify}

Tables \ref{tab:facts_breakdown_zs}, \ref{tab:facts_breakdown_hm}, \ref{tab:facts_breakdown_cite_hm}, \ref{tab:facts_breakdown_ext_sup} and \ref{tab:facts_breakdown_ext_repl} respectively show our manual annotations for  the Zero-shot summary, the original HMerge summary and its adaptation with citations Cite-HMerge, the Extract-Support  and  Extract-Replace summaries. "\greencheck" means the statement is supported by the source book text. "\redcross" means the statement or parts thereof are contradicted by the source book text. "\textbf{--}" means that the statement or parts thereof are not present in the source book text. We implemented the following annotation procedure:

\begin{itemize}
    \item If every part of the statement is  entailed by the book source, we  classify the statement as Correct ("\greencheck").
    \item Otherwise, we check whether some parts of the statement  contradict the book source and  classify the statement as Incorrect ("\redcross"). A piece of text  \textit{contradicts} the source if it has information is incompatible or contradictory with explicit statements or facts presented in the book source.
    \item If none of the above apply, we classify the statement as Not Present ("\textbf{---}"), since it expresses information that is neither directly supported or contradicted by the source (and as a result we cannot make a definitive judgment about its validity). 
\end{itemize}

\begin{table*}
    \centering
    \renewcommand{\arraystretch}{1.2}
    \fontsize{9}{10}\selectfont
    \begin{tabular}{p{14cm}|c}
    \hline
    Lina Solavita is an insurance investigator who moves to Knockemout, a small town in northern Virginia. & \greencheck \\
    \hline
    Lina is in Knockemout to find a 1948 Porsche 356 convertible that was stolen from a client. & \greencheck \\
    \hline
    Lina is staying in an apartment next to Nash Morgan, who is the chief of police in Knockemout. & \greencheck \\
    \hline
    Nash Morgan is recovering from a shooting that left him with bullet wounds and memory loss. & \greencheck \\
    \hline
    Nash Morgan struggles to cope with his new reality and is haunted by the sound associated with the moment he was shot. & \greencheck \\
    \hline
    Lina is initially attracted to Nash Morgan, but their relationship is complicated by his vulnerability and her job. & \redcross \\
    \hline
    Lina's job as an insurance investigator requires her to keep secrets. & \textbf{---} \\
    \hline
    Lina is dealing with her own emotional baggage, including a heart condition that almost killed her at the age of 15. & \greencheck \\
    \hline
    A U.S. marshal named Nolan Graham is watching Nash Morgan due to a threat against him. & \greencheck \\
    \hline
    Lina is caught in the middle of the situation between Nash Morgan and Nolan Graham. & \greencheck \\
    \hline
    As Lina and Nash spend time together, they develop a strong connection. & \greencheck \\
    \hline
    Lina and Nash's differences and secrets threaten to tear their relationship apart. & \redcross \\
    \hline
    Lina's past, including her history with Nolan Graham and her job, is revealed gradually throughout the story. & \greencheck \\
    \hline
    Nash Morgan feels torn between his growing feelings for Lina and his need for honesty and trust. & \textbf{---} \\
    \hline
    Lina's investigation leads her into danger, prompting Nash to confront his vulnerabilities. & \textbf{---} \\
    \hline
    Lina and Nash are eventually kidnapped by a group of bikers working for Grim, a motorcycle club leader. & \redcross \\
    \hline
    Grim reveals that Duncan Hugo, the man who shot Nash, is still in town and planning to take over his father's criminal empire. & \greencheck \\
    \hline
    The relationship between Lina and Nash deepens as they work together to uncover Duncan Hugo's plans. & \greencheck \\
    \hline
    Lina and Nash face challenges from Tate Dilton, a disgraced police officer, and Lucian Rollins, who is Nash's friend. & \greencheck \\
    \hline
    The story includes a dramatic confrontation at a library's Book or Treat event where Lina and her friends are attacked by bikers. & \redcross \\
    \hline
    Lina and Nash's relationship is a central theme of the story, marked by undeniable chemistry. & \greencheck \\
    \hline
    Secrets and lies threaten to destroy the trust between Lina and Nash, potentially ending their relationship. & \redcross \\
    \hline
    The story is a complex exploration of love, trust, and vulnerability set in a small town with its own secrets. & \textbf{---} \\
    \hline
    The cast of characters includes Mrs. Tweedy, a feisty elderly woman, and Naomi, a kind-hearted community outreach coordinator who becomes Lina's friend. & \greencheck \\
    \hline
    The narrative is described as a gripping and emotional ride with a complex plot that keeps readers guessing until the end. & \textbf{---} \\
    \hline
    \end{tabular}
    \caption{Fact annotations for the Zero-shot summary.}
    \label{tab:facts_breakdown_zs}
\end{table*}

\begin{table*}
    \centering
    \renewcommand{\arraystretch}{1.2}
    \fontsize{8}{9}\selectfont
    \begin{tabular}{p{14cm}|c}
    \hline
    Lina Solavita is an investigator for Pritzger Insurance, and she moved to Knockemout, Virginia. & \greencheck \\
    \hline
    Nash Morgan is the police chief of Knockemout, Virginia, who was shot in the line of duty and was struggling to accept his new reality. & \greencheck \\
    \hline
    Lina's presence in Knockemout sparked a confrontation with US Marshal Nolan Graham, who was shadowing Nash. & \redcross \\
    \hline
    Lina has a past marked by a medical condition and a complicated relationship with physical touch, which made her interactions with Nash awkward. & \redcross \\
    \hline
    Lina received a call from her mother, Bonnie, who was checking in on her while she was settling into her new apartment. & \greencheck \\
    \hline
    Lina reassured her mother that she was fine and was enjoying her time in Knockemout, but Bonnie was not convinced and planned to visit soon. & \greencheck \\
    \hline
    Knox Morgan, Nash's brother, knocked on Lina's door to talk to Nash about a matter that he found important. & \redcross \\
    \hline
    Knox insisted on discussing something with Nash, even though Nash was not interested in talking. & \textbf{---} \\
    \hline
    As Lina waited for Nash and Knox's conversation to conclude, she was left wondering about the relationship between the two brothers. & \textbf{---} \\
    \hline
    Despite her challenges, Lina was determined to make the most of her time in Knockemout. & \greencheck \\
    \hline
    Lina started to feel a connection with Nash and hoped to help him heal and find his way again. & \textbf{---} \\
    \hline
    Nash was struggling with his feelings for Lina while haunted by nightmares and trying to heal from his past. & \redcross \\
    \hline
    Lina is portrayed as a strong yet vulnerable and guarded character in the story. & \greencheck \\
    \hline
    Nash is depicted as a complex character with a troubled past who is also kind and heroic. & \greencheck \\
    \hline
    The chemistry between Lina and Nash is described as palpable, with their interactions filled with tension and humor. & \textbf{---} \\
    \hline
    As Lina navigated her feelings for Nash, she confronted her own vulnerabilities and learned to trust him. & \greencheck \\
    \hline
    Nash's brother, Knox, was a constant presence in the background, providing advice and guidance to him. & \greencheck \\
    \hline
    Nash was determined to make his own decisions and forge his own path despite his brother's influence. & \textbf{---} \\
    \hline
    The investigation into police officer misconduct and the activities of mysterious henchmen was ongoing in Knockemout. & \greencheck \\
    \hline
    Lina's friends, Stef, Naomi, and Sloane, were supporting her while dealing with their own personal issues, including a recent abduction. & \redcross \\
    \hline
    Nolan Graham, a US Marshal, was trying to get Lina to open up about her past and was acting as a supportive friend. & \redcross \\
    \hline
    Lina and Nolan had a non-date drink together, during which Lina began to open up about her feelings and experiences. & \redcross \\
    \hline
    The relationship between Nash and Lina was complex and multifaceted, shaped by their past experiences. & \greencheck \\
    \hline
    Lina was kidnapped by someone known as Cereal Aisle Guy, who was revealed to be Nikos. & \greencheck \\
    \hline
    Lina was taken to a barn-like building where she met Tate Dilton, a corrupt cop, and Duncan Hugo, the mastermind behind the kidnapping. & \greencheck \\
    \hline
    Duncan Hugo revealed that he hired Tate Dilton to shoot Nash, but Dilton shot him in cold blood on the highway instead. & \greencheck \\
    \hline
    Duncan's plan was to use Lena to lure Nash to the barn, where Nash would be killed. & \redcross \\
    \hline
    While held captive, Lina learned about Duncan’s goal to take over his father's business and Tate’s personal vendetta against Nash. & \greencheck \\
    \hline
    Nash and his friends were searching for Lina by following a lead on a partial plate number that Waylay memorized. & \greencheck \\
    \hline
    Waylay, a character in the story, helped Lina by recognizing her name and informed Nash of her situation during an online game. & \redcross \\
    \hline
    Nash tracked down a lead to a Ford Fusion belonging to Mark Nikos, as he searched for Lina. & \greencheck \\
    \hline
    Nash, along with Knox and Nolan, approached the farm where Lina was being held by Tate and Duncan. & \greencheck \\
    \hline
    A shootout occurred at the farm during the rescue attempt, where Lina managed to use a pressure washer to defend herself. & \redcross \\
    \hline
    Lina stabbed Duncan Hugo with a pitchfork and escaped to a Porsche, where she met up with Knox and Nolan. & \redcross \\
    \hline
    During the escape attempt, Dilton shot Nolan, and Lina and Knox sought cover behind a tractor. & \greencheck \\
    \hline
    Nash proposed to Lina, and she accepted his proposal amid the ongoing chaos. & \greencheck \\
    \hline
    Knox managed to take Lina and Nolan to safety while Nash stayed behind to confront Tate Dilton. & \greencheck \\
    \hline
    Nash successfully killed Tate Dilton by shooting him three times after outmaneuvering him. & \greencheck \\
    \hline
    The story features a wedding ceremony where Knox marries Naomi, with Lina in attendance and Nash serving as a groomsman. & \redcross \\
    \hline
    After the wedding ceremony, Nash and Lina shared a dance, during which he expressed thoughts about their future together. & \greencheck \\
    \hline
    Nash revealed to Lina that he got a tattoo of angel wings on his butt, which she found romantic. & \greencheck \\
    \hline
    During their dance, Nash's father, Duke, congratulated them on their engagement and seemed to be making amends for past mistakes. & \greencheck \\
    \hline
    Nash and Lina sneaked away to Nash's old bedroom, where they shared a romantic moment that was interrupted by the unexpected arrival of Lina's parents. & \greencheck \\
    \hline
    Lina felt relieved that her parents had finally met Nash, even though he was caught off guard by their unexpected presence. & \textbf{---} \\
    \hline
    \end{tabular}
    \caption{Fact annotations for the original HMerge summary.}
    \label{tab:facts_breakdown_hm}
\end{table*}

\begin{table*}
    \centering
    \renewcommand{\arraystretch}{1.2}
    \fontsize{9}{10}\selectfont
    \begin{tabular}{p{14cm}|c}
    \hline
    Nash Morgan is a police chief who is struggling to cope with the emotional aftermath of being shot in the line of duty. His emotional numbness is affecting his relationships with those around him. & \greencheck \\
    \hline
    Angelina Solavita, also known as Lina, is Nash's new neighbor and an insurance agent. Lina is also Nash's brother's ex-girlfriend from college. & \redcross \\
    \hline
    Lina is investigating a case involving a stolen car, which leads her to cross paths with Nash. Nash is still recovering from his injuries during their encounters. & \greencheck \\
    \hline
    As Nash and Lina spend more time together, they grow closer, but their relationship is complicated by Nash's past trauma and Lina's secrets. Nash's emotional state is affected by his experiences as a police officer. & \greencheck \\
    \hline
    Lina is searching for a car that Duncan Hugo stole and is looking for leads in the crime scene files. She feels frustrated that she is not being included in the plans to deal with Hugo. & \redcross \\
    \hline
    Lina observes a fight between two brothers, Wendell Baker and his brother, and calls 911. Before the police arrive, she is abducted by two people in a van. & \greencheck \\
    \hline
    The two abductors take Lina to meet Grim, the leader of a motorcycle club. Grim has information about Duncan Hugo, who is still in town and planning to take over his father's crime business. & \greencheck \\
    \hline
    Nash is concerned about Lina's safety and believes she should not be involved in the investigation regarding Hugo. He is pursuing her to ensure her well-being. & \greencheck \\
    \hline
    Lina attends a Halloween party at the library and runs into an old friend from high school, Angie. Angie is coping with her son Austin’s recovery from leukemia and reflects on how to be a better parent. & \greencheck \\
    \hline
    Lina reconnects with Angie, who apologizes for not being a better friend. Lina reflects on how she pushed her friends away after her own health issues. & \greencheck \\
    \hline
    Nash and his friends are preparing for a confrontation with Duncan Hugo and his men. Lina is trying to stay one step ahead of Hugo’s plans. & \textbf{---} \\
    \hline
    Lina and Nash's relationship is complicated by her feelings and past experiences. She is dealing with her vulnerabilities while navigating their connection. & \greencheck \\
    \hline
    Angelina, dressed as Nancy Drew, is approached at the Halloween party by Nash Morgan and his brothers, Knox, Nolan, and Lucian. Their group is interrupted by a drunk man grabbing Sloane. & \redcross \\
    \hline
    Nash arrests the drunk man and his friend, who had also fought with Nolan. After the altercation, Nash asks Angelina to dance, which she initially hesitates to do. & \redcross \\
    \hline
    While dancing, the tension between Nash and Angelina becomes palpable, leading Nash to pull her away to a quiet room for a passionate kiss. & \textbf{---} \\
    \hline
    Angelina reflects on her life-changing experience with Nash, which leaves her questioning her feelings. The night before, they had an intimate encounter that overwhelmed her. & \greencheck \\
    \hline
    Nash and Angelina made a deal to work together and be open with each other after their intense encounter. Nash realizes he has been given a second chance with Angelina. & \greencheck \\
    \hline
    Nash sends Angelina a playful text message after their encounter, teasing her about her exhaustion. They banter and agree to go out to dinner together. & \textbf{---} \\
    \hline
    Nash speaks with Sloane, who discovers that he and Angelina had slept together. Their dinner date ends on a romantic note, enhancing the connection between them. & \textbf{---} \\
    \hline
    Angelina expresses her desire to help catch Duncan Hugo and keep Nash safe. Nash is touched by her concern and realizes that Angelina cares about him deeply. & \greencheck \\
    \hline
    \end{tabular}
    \caption{Fact annotations for the Cite-HMerge summary.}
    \label{tab:facts_breakdown_cite_hm}
\end{table*}

\begin{table*}
    \centering
    \renewcommand{\arraystretch}{1.2}
    \fontsize{9}{10}\selectfont
    \begin{tabular}{p{14cm}|c}
    \hline
    Police Chief Nash Morgan struggles to cope with the aftermath of being shot in the line of duty in the small town of Knockemout. & \greencheck \\
    \hline
    Nash is emotionally numb and feeling useless while going through the motions of his job without passion or interest. & \greencheck \\
    \hline
    Angelina "Lina" Solavita, Nash's brother's ex-girlfriend, moves into the apartment next door to Nash. & \greencheck \\
    \hline
    Lina is a beautiful and charming woman who works as an insurance investigator. & \greencheck \\
    \hline
    Nash is drawn to Lina despite his emotional numbness. & \greencheck \\
    \hline
    Lina is in Knockemout investigating a case involving law enforcement officers and informants, including Nash, who was targeted by Duncan Hugo. & \redcross \\
    \hline
    Duncan Hugo is the son of a wealthy and influential man who presents a threat to Nash and others. & \greencheck \\
    \hline
    Lina's investigation includes searching for a stolen 1948 Porsche 356 convertible that is priceless to her client. & \greencheck \\
    \hline
    Lina befriends Naomi, Knox's fiancée, and Sloane, a librarian, while navigating her investigation. & \greencheck \\
    \hline
    Nash is recovering from his injuries and is experiencing memory loss and panic attacks. & \greencheck \\
    \hline
    Nash is haunted by flashbacks of the shooting incident. & \greencheck \\
    \hline
    Lina acts as a calm and soothing presence in Nash's life, helping him feel more at ease. & \textbf{---} \\
    \hline
    Nash and Lina confront secrets and mysteries surrounding them as their attraction grows. & \greencheck \\
    \hline
    Lina observes a confrontation between Wendell Baker and his brother at a row home during her investigation. & \greencheck \\
    \hline
    Lina is abducted by a group of people in ski masks during the confrontation, who turn out to be associates of a man named Grim. & \greencheck \\
    \hline
    Grim is the leader of a motorcycle club who possesses information about Duncan Hugo's whereabouts. & \greencheck \\
    \hline
    As Nash and Lina's relationship deepens, Lina begins to confront her fears about opening up to Nash and her parents. & \textbf{---} \\
    \hline
    The investigation into Duncan Hugo's whereabouts continues as a team works together to gather information and ensure their safety. & \greencheck \\
    \hline
    The team plans to use Nash as bait to lure Duncan Hugo out of hiding. & \redcross \\
    \hline
    Lucian Rollins, a wealthy and influential man, expresses concern over Lina's intentions and warns her to stay away from Nash. & \redcross \\
    \hline
    Lina confronts Lucian for trying to manipulate her and warns him to respect Nash's boundaries. & \greencheck \\
    \hline
    Despite initial tension, Lucian becomes an ally to Nash and Lina, assisting them in their quest to bring Hugo to justice. & \greencheck \\
    \hline
    As the team prepares for a showdown with Duncan Hugo, tensions rise regarding the safety of Naomi and her daughter, Waylay. & \textbf{---} \\
    \hline
    The team is determined to bring Duncan Hugo to justice despite their concerns. & \greencheck \\
    \hline
    Nash and Lina's relationship faces tests as they work together to confront Duncan Hugo. & \greencheck \\
    \hline
    Nash proposes to Lina amidst chaos, and she accepts his proposal. & \greencheck \\
    \hline
    The story concludes with Nash and Lina surrounded by police and responders, emerging from an intense situation as a new day begins. & \redcross \\
    \hline
    Nash struggles with feelings of guilt and helplessness after a traumatic incident where a woman was severely injured in a car accident. & \greencheck \\
    \hline
    The memory of his mother's death in a similar accident adds to Nash's guilt. & \redcross \\
    \hline
    Nash pushes away Lina, telling her he needs space and does not want her around anymore. & \greencheck \\
    \hline
    Despite Nash's rejection, Lina comforts him, reminding him that he is a hero for facing his demons. & \greencheck \\
    \hline
    The story explores Nash's complicated relationship with his father, Duke, who has been absent but is trying to reconnect with his sons. & \greencheck \\
    \hline
    Nash feels wary but ultimately happy to see his father making progress and is grateful for his presence at Knox's wedding. & \redcross \\
    \hline
    \end{tabular}
    \caption{Fact annotations for the Extract-Support summary.}
    \label{tab:facts_breakdown_ext_sup}
\end{table*}

\begin{table*}
    \centering
    \renewcommand{\arraystretch}{1.2}
    \fontsize{9}{10}\selectfont
    \begin{tabular}{p{14cm}|c}
        \hline
        Sloane is a former sophomore who bought her house from her parents when they moved. & \greencheck \\
        \hline
        Lucian is a former senior who kept his mom's house. & \greencheck \\
        \hline
        Lou and Amanda volunteered to walk Liza J home, which sparked a discussion about who would foster her. & \textbf{---} \\
        \hline
        Anthony Hugo is a notorious crime lord who controls a vast territory that includes Washington, DC, and Baltimore. & \greencheck \\
        \hline
        Naomi and Sloane simultaneously expressed interest in fostering Liza J. & \textbf{---} \\
        \hline
        Knox and Lucian stood alongside Nash and Nolan while Officer Dilton revealed that Mr. Peters was Knockemout's solicitor. & \redcross \\
        \hline
        Lucian proposed a theory that Anthony Hugo had helped his son leave the country. & \greencheck \\
        \hline
        The community organized a party with library patrons buying tickets and business sponsors donating food and drinks, persuaded by Sloane. & \greencheck \\
        \hline
        The library retained the profits from the event. & \greencheck \\
        \hline
        Special Agent Idler has a personal vendetta against Anthony Hugo and is determined to bring him to justice. & \greencheck \\
        \hline
        Naomi and Knox, along with their extended families, helped maintain order in the community. & \greencheck \\
        \hline
        Nash Morgan posed a significant threat to women everywhere. & \redcross \\
        \hline
        Naomi was reminded to take a time-out as she, Sloane, and Mrs. Tweedy approached her soon-to-be husband. & \textbf{---} \\
        \hline
        Tallulah and Justice St. John, along with Gael and his husband Isaac, occupied a table enjoying their monthly double date. & \greencheck \\
        \hline
        Naomi and Sloane were taking their assignments seriously, and Sloane raised her hand to contribute to the discussion. & \textbf{---} \\
        \hline
        Knox, Nolan, Lucian, and others gathered in Knox's secret lair office according to the Knockemout grapevine. & \redcross \\
        \hline
        Knox shared a long, meaningful look with his brother. & \redcross \\
        \hline
        Jeremy Trent, the former captain of the baseball team, beat Dilton for homecoming king during their senior year. & \greencheck \\
        \hline
        Knox drove Waylay and Liza J home to Naomi, who was tracked as being at home. & \greencheck \\
        \hline
        Naomi and Sloane approached others with a proposal about their new venture. & \textbf{---} \\
        \hline
        Naomi's wedding was scheduled for the next day while Waylay was just a child. & \greencheck \\
        \hline
        Lina was spotted in the photo booth with Sloane and Fi as they celebrated the upcoming wedding. & \greencheck \\
        \hline
        Discussions about real estate, marriage, and babies filled the atmosphere. & \redcross \\
        \hline
        Liza J's presence reminded the community of the challenges that lay ahead. & \redcross \\
        \hline
        Lucian and a private security team were tasked with ensuring the safety of Naomi, Sloane, and others. & \greencheck \\
        \hline
        Lucian was seen clutching his prized Smith \& Wesson six-shooter, symbolizing his commitment to protection. & \redcross \\
        \hline
        Lucian nodded and followed Melissa inside to address the looming threat. & \redcross \\
        \hline
        Duncan Hugo was wanted for attempted murder of a law enforcement officer, domestic violence, and assault. & \greencheck \\
        \hline
        Concerns about Naomi's safety were raised, and Lucian reassured everyone that she was being protected. & \textbf{---} \\
        \hline
        Duncan Hugo's careless attitude contrasted sharply with the gravity of the situation. & \textbf{---} \\
        \hline
        A warning was issued to call Nash and inform him that Duncan Hugo was sending Tate Dilton to their house. & \textbf{---} \\
        \hline
        The situation was deemed dire, necessitating immediate action. & \greencheck \\
        \hline
        Explaining the situation to a twelve-year-old was seen as daunting. & \greencheck \\
        \hline
        Duncan Hugo was in custody, but the threat to safety remained. & \greencheck \\
        \hline
        A plan was devised to fly in Liza J's family to surprise her as a gesture of joy amidst chaos. & \textbf{---} \\
        \hline
        Lou recalled a family tradition of grand gestures, like surprising Mandy with a three-week cruise. & \greencheck \\
        \hline
        Knox, the groom, was the focus of attention, and it was essential to ensure his safety along with Nolan and Liza J. & \textbf{---} \\
        \hline
        Jeremiah and Lou asked unanimously if they should take Knox and others to a safe location. & \textbf{---} \\
        \hline
        The atmosphere was tense with a sound of a Ford Fusion
        belonging to Mark Nikos, reminding everyone of the outside
        world. & \textbf{---} \\
        \hline
        Community support was evident as events unfolded. & \greencheck \\
        \hline
        Naomi and Sloane's commitment to their assignments showed dedication to the community. & \greencheck \\
        \hline
        The presence of Lucian and the private security team provided reassurance against threats posed by Duncan Hugo and Nash Morgan. & \redcross \\
        \hline
        The celebration of Naomi's wedding symbolized hope and community unity in the face of adversity. & \redcross \\
        \hline
    \end{tabular}
    \caption{Fact annotations for the Extract-Replace summary.}
    \label{tab:facts_breakdown_ext_repl}
\end{table*}

\end{document}